\documentclass[sigconf]{acmart}
\usepackage{algorithm}
\usepackage{algorithmic}
\usepackage{subfigure}
\usepackage{libertine}

\AtBeginDocument{%
  \providecommand\BibTeX{{%
    \normalfont B\kern-0.5em{\scshape i\kern-0.25em b}\kern-0.8em\TeX}}}

\copyrightyear{2020}
\acmYear{2020}
\setcopyright{acmcopyright}\acmConference[CIKM '20]{Proceedings of the 29th ACM
International Conference on Information and Knowledge Management}{October
19--23, 2020}{Virtual Event, Ireland}
\acmBooktitle{Proceedings of the 29th ACM International Conference on Information
and Knowledge Management (CIKM '20), October 19--23, 2020, Virtual Event, Ireland}
\acmPrice{15.00}
\acmDOI{10.1145/3340531.3411865}
\acmISBN{978-1-4503-6859-9/20/10}

\begin{document}
\fancyhead{}

\title{VN Network: Embedding Newly Emerging Entities with Virtual Neighbors}

\author{Yongquan He}
\email{heyongquan@iie.ac.cn}
\affiliation{%
  \institution{School of Cyber Security, University of Chinese Academy of Sciences\\
Institute of Information Engineering, Chinese Academy of Sciences}
}

\author{Zihan Wang}
\email{zihanwang.sdu@gmail.com}
\affiliation{%
  \institution{School of Computer Science and Technology, Shandong University}
}

\author{Peng Zhang}
\authornote{Corresponding author}
\email{pengzhang@iie.ac.cn}
\affiliation{%
  \institution{Institute of Information Engineering, Chinese Academy of Sciences}
}

\author{Zhaopeng Tu}
\email{tuzhaopeng@gmail.com}
\affiliation{%
  \institution{Tencent AI Lab}
}

\author{Zhaochun Ren}
\email{zhaochun.ren@sdu.edu.cn}
\affiliation{%
  \institution{Shandong University}
}

\begin{abstract}
  Embedding entities and relations into continuous vector spaces has attracted a surge of interest in recent years.
  Most embedding methods assume that all test entities are available during training,
  which makes it time-consuming to retrain embeddings for newly emerging entities.
  To address this issue, recent works apply the graph neural network on the existing neighbors of the unseen entities.
  In this paper, we propose a novel framework, namely Virtual Neighbor (VN) network, to address three key challenges.
  Firstly, to reduce the \textbf{neighbor sparsity problem}, we introduce the concept of the virtual neighbors inferred by rules.
  And we assign soft labels to these neighbors by solving a rule-constrained problem,
rather than simply regarding them as unquestionably true.
  Secondly, many existing methods only use one-hop or two-hop neighbors for aggregation and ignore the distant information that may be helpful.
  Instead, we identify both logic and symmetric path rules to capture \textbf{complex patterns}.
  Finally, instead of one-time injection of rules,
  we employ an iterative learning scheme between the embedding method and virtual neighbor prediction to \textbf{capture the interactions within}.
  Experimental results on two knowledge graph completion tasks demonstrate that our VN network significantly outperforms state-of-the-art baselines. 
  Furthermore, results on Subject/Object-R show that our proposed VN network is highly robust to the neighbor sparsity problem.
\end{abstract}

\maketitle

\section{Introduction}

Recently, knowledge graphs (KGs) such as Freebase \cite{DBLP:conf/sigmod/BollackerEPST08} and Google's Knowledge Vault \cite{DBLP:conf/kdd/0001GHHLMSSZ14} 
have proven to be extremely useful resources for many natural language processing(NLP) relevant applications \cite{DBLP:conf/emnlp/PritskerCM15,DBLP:conf/acl/YangM17}.
A typical KG can be considered as a multi-relational graph, where nodes and various types of edges reflect entities and relations, respectively.
Each edge is presented as a triple of the form (\emph{head entity}, \emph{relation}, \emph{tail entity}), e.g., (\emph{John}, \emph{bornedIn}, \emph{Athens}).
Although effective in representing structured data, the symbolic nature of triples often makes KGs hard to manipulate, and most of the KGs are far from complete compared to existing facts in the real world.
To tackle this issue, KG embedding methods have been proposed, aiming to embed entities (nodes) and relations (edges) into the continuous vector spaces.
In that way, such kind of methods can encode the structure of the KGs and simplify the manipulation.
And KG embeddings contain rich semantic information and can facilitate KG completion and inference \cite{DBLP:conf/icml/NickelTK11,DBLP:conf/aaai/GuoWWWG18,DBLP:conf/acl/WangWGD18,DBLP:conf/ijcai/WangRHZH19}.

Despite the success of embedding methods, previous approaches still can not handle newly emerging entities as all the entities need to be seen in the training process.
In that case, newly emerging entities are unknown to the original system.
Without retraining on the whole KG, the original system is difficult to predict missing facts about new entities.
As knowledge graphs evolve dynamically, new entities appear along with emerging events or new products \cite{DBLP:conf/ijcai/HamaguchiOSM17,DBLP:conf/aaai/ShiW18}.
Therefore, an inductive learning framework is needed for avoiding a time-consuming retraining process.
The key idea of an inductive learning framework is to generalize the original system to the newly observed subgraph,
which requires the recognition of the neighborhood structural properties of the unseen nodes \cite{DBLP:conf/nips/HamiltonYL17}.
MEAN \cite{DBLP:conf/ijcai/HamaguchiOSM17} applies the graph neural network (GNN) on both the observed and unseen entities,
mapping the newly emerging entities into embeddings by aggregating their known neighbors.
However, it neglects the different edge types (relations) and adopts the simple mean pooling aggregator.
Based on that, logic attention based neighborhood aggregation network (LAN) \cite{DBLP:conf/aaai/WangHLP19} is proposed, 
which employs both logic rules and neural network attention mechanisms to capture the redundancy information and concentrate on the query relevant facts. 

\begin{figure}[h]
  \centering
  \includegraphics[width=2.6in]{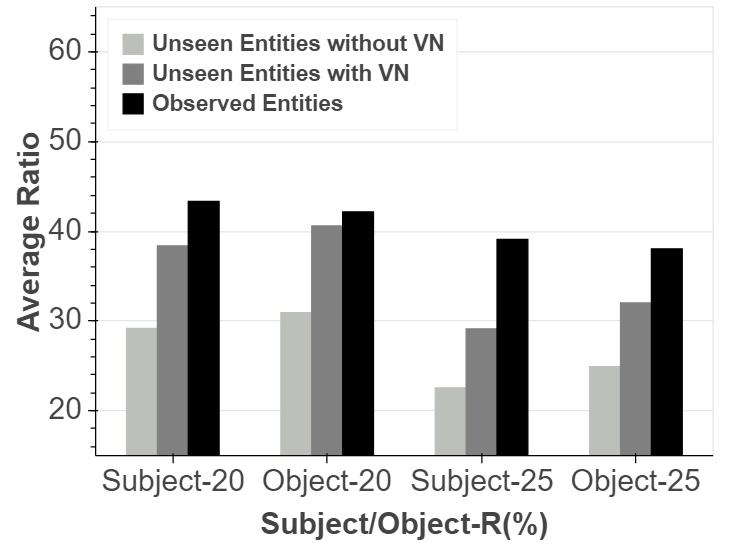}
  \caption{The average ratio of neighbors to predicted facts with and without virtual neighbors.}
  \Description{The average ratio of neighbors to predicted facts with and without virtual neighbors.}
  \label{fig:neighbor}
\end{figure}

Although the above methods can solve the unseen entity problem to some extent,
there are still several problems not fully addressed.

\textbf{Neighbor Sparsity Problem:}
In fact, compared to observed entities, we generally do not have enough information to learn good representations for newly emerging entities.
As shown in Figure~\ref{fig:neighbor},
in the subject-\{20, 25\} and object-\{20, 25\} datasets \cite{DBLP:conf/aaai/WangHLP19} which obtained by selecting 20\% or 25\% triples from FB15K \cite{DBLP:conf/nips/BordesUGWY13} test set, 
to sample the candidate unseen entities.
And then use these candidate unseen entities to split the original training set, to ensure that these unseen entities are not observed during training process.
We can see that the average ratio of the known neighbors to the predicted triples for the unseen entities is lower than the observed entities by at least 11 in the current graph,
which makes the knowledge representation learning much more difficult when collecting neighborhood information.
For example, in the subject-20, we only know average 29.19 triples of facts about the unseen entity, but for the observed entities is 43.32.
Both methods above suffer this sparsity problem.

\begin{figure}[h]
  \centering
  \includegraphics[width=3.0in]{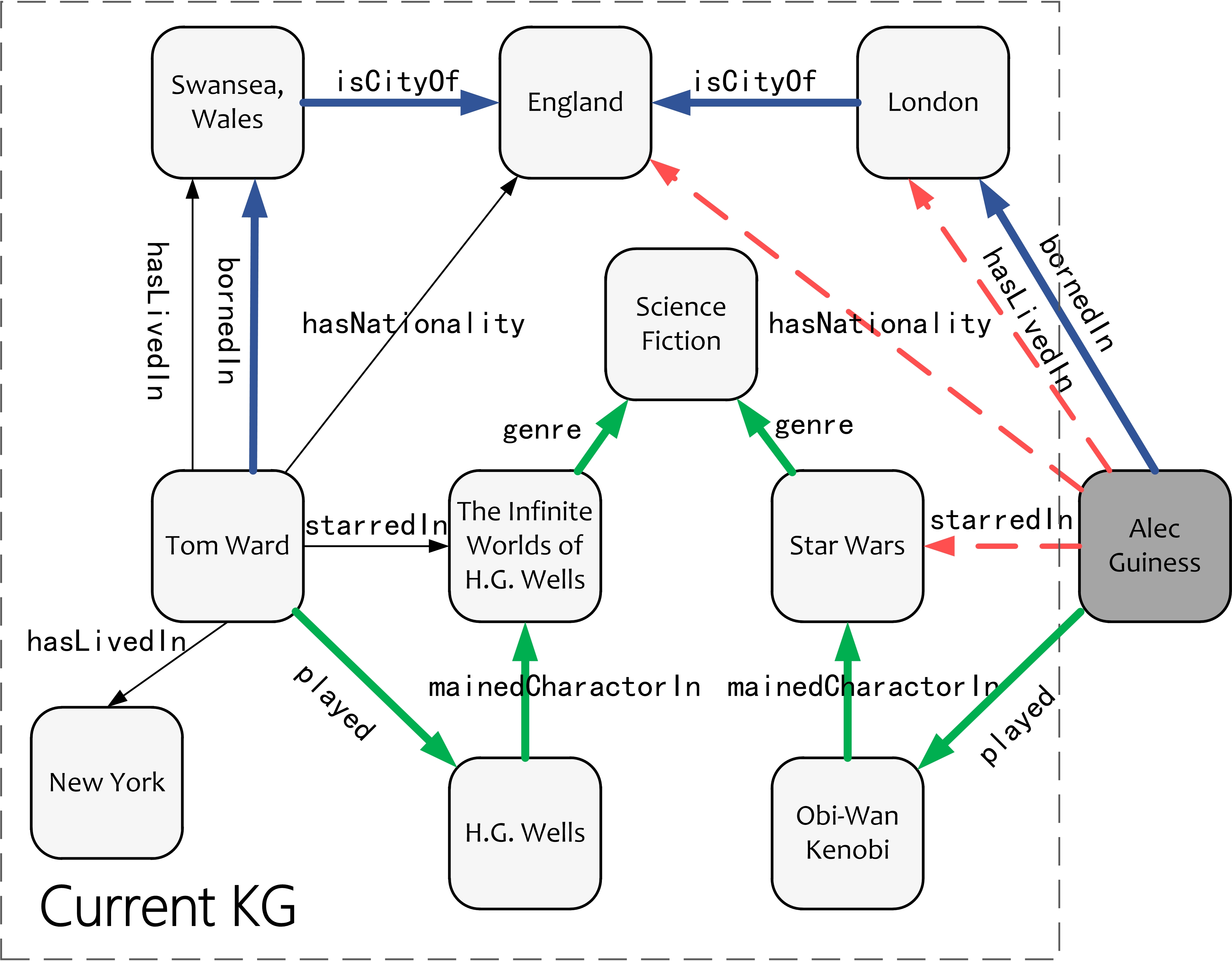}
  \caption{The example of the unseen entities problem about original KG.}
  \Description{The example of the unseen entities problem about original KG.}
  \label{fig:examplesp}
\end{figure}

\textbf{Complex Pattern:}
The previous methods mainly focus on the 1 or 2 hops of the connecting structures, or the 1 hop rules about the relation dependence, 
while there are many other long-distance dependencies that are helpful for the knowledge base completion. 
As Figure~\ref{fig:examplesp} describes, 
at test time, we receive new triples containing the entity "\emph{Alec Guiness}", which is not observed in the original KG.
From the first path (green) we can find the entity "\emph{Tom}", which has similar semantic about actors' roles.
We know the fact that "\emph{Tom}" was in the science fiction "\emph{The Infinite Worlds}", 
and such fact can help complete the missing edge "\emph{starredIn}",
which means an actor played a character who belongs to a science fiction, he must be in that science fiction.
And the second path (blue) hints that they were born in different cities in the same country.
We can analogy that "\emph{Alec}" had lived in England by finding the pattern about "\emph{Tom}",
meaning that a person was born in a city of a country, he must have lived in that country.
So we can obtain more information and capture long dependencies between entities through these long-distance paths to assist with short logic rules.

\textbf{Interactions Between the Rule Inference and Embedding Learning:}
LAN regards the confidence level of the logic rules as the constant weights for the aggregator,
while embeddings encoded with rich semantics can tune the results inferred by rules.
As RUGE \cite{DBLP:conf/aaai/GuoWWWG18} and IterE \cite{zhang2019iteratively} mentioned, rules can infer new facts more accurately with the refined embeddings, 
and newly inferred facts can also help to learn the embeddings of higher quality.

In this work, we propose a novel inductive learning framework, VN network, to embed newly emerging entities.
The VN network consists of three main components, including the virtual neighbor prediction, encoder, and decoder.
To reduce the sparsity problem, we introduce the concept of the virtual neighbors and employ the rule-based virtual neighbor prediction algorithm.
Meanwhile, to capture more information in the KGs, we adopt the logic rules and identify the long-distance symmetric path rules.
In that case, the soft labels of triples inferred by rules are computed by solving a convex rule-constrained problem over the current KG embeddings.
Then, the KG with the predicted virtual neighbors is inputted to the GNN-based encoder composed of several structure aware layers and one query aware layer.
Finally, we output the embeddings from the encoder to the decoder for predicting the missing facts.
At each iteration, to combine with rules in an iterative manner, we first refine the soft labels using the current embeddings and then obtain the optimal current KG embeddings by minimizing the global loss over both the hard labeled and soft labeled triples.

Our contributions are summarized as follows:
1)We propose a novel inductive learning framework to handle the unseen entities.
2)We develop a virtual neighbor prediction method to reduce the sparsity problem, 
identify the logic and symmetric path rules to capture more information and establish an iterative refinement scheme over the soft labels and KG embeddings.
3)We conduct two types of the knowledge graph completion tasks on WordNet11, FB15K and YAGO37 to demonstrate the effectiveness of the VN network,
and show about a significant improvement over the state-of-the-art LAN.
\section{Related Work}

In recent years, we have witnessed increasing interests in embedding methods for KGs.
Such methods have become one of the most important techniques on knowledge base completion (KBC), aiming to map the entities and relations into continuous vector space.
Recent works can be mainly classified into two categories: 
some of them proposed more complicated scoring functions and deeper frameworks, such as TransE extensions \cite{DBLP:conf/aaai/WangZFC14,DBLP:conf/acl/JiHXL015}, 
RESCAL extensions \cite{yang2014embedding,DBLP:conf/icml/TrouillonWRGB16},
and combining with deep neural network \cite{DBLP:conf/ijcai/WangRHZH19,DBLP:conf/aaai/ShangTHBHZ19}.
Others tried to further incorporate other information available, including relation paths \cite{DBLP:conf/acl/NeelakantanRM15,xie2016representation,xiao2017ssp}, typing information \cite{zhang2018knowledge}, visual information \cite{DBLP:conf/ijcai/XieLLS17} 
and logic rules \cite{DBLP:conf/aaai/GuoWWWG18,DBLP:conf/acl/WangWGD18,zhang2019iteratively}.
More detailed reviews can be found in \cite{ji2020survey} and \cite{DBLP:journals/tkde/WangMWG17}.

Although proving the success in the KBC task, traditional KG embedding methods still fall short on the newly emerging entities issue.
Previous embedding frameworks require all the entities should be seen during the training process. 
However, KG evolves frequently, and a large number of new entities emerge almost on a daily basis, especially between late 2015 and early 2016 \cite{DBLP:conf/aaai/ShiW18}.
Meanwhile, retraining on the whole KG to obtain embeddings of the new entities is extremely time-consuming. 
To address the newly emerging entities problem, several works utilize other types of information, such as text descriptions and image \cite{DBLP:conf/ijcai/XieLLS17,DBLP:conf/aaai/ShiW18}, to predict new facts.
These methods are still limited when text and image information are not sufficient or provided.
Moreover, we also need knowledge which are usually graph-specific under domain expert guidance to use these information.
Compared with above information, rules are easier to obtain,
and there are many efficient rule mining methods for knowledge graph, such as AMIE+ \cite{DBLP:journals/vldb/GalarragaTHS15} or RLvLR \cite{omran2019embedding}.
And rules are inherently inductive since they are independent of node identities,
which can assist embedding learning as useful information for reasoning.
The efficiency of learning embedding and rules in an iterative way has proved \cite{DBLP:conf/aaai/GuoWWWG18,zhang2019iteratively}.
But most general rule-based methods only focus on one-hop and two-hop relations.
In this paper, through introducing the symmetric path rules, we can capture long-distance dependencies between entities.

There are some other works such as DKGE \cite{wu2019efficiently} aiming at solving the emerging facts about KGs, but the most relevant works to ours are LAN \cite{DBLP:conf/ijcai/HamaguchiOSM17} and MEAN \cite{DBLP:conf/aaai/WangHLP19}, which focus on the representation learning of the unseen entities by aggregating neighbors.
MEAN \cite{DBLP:conf/ijcai/HamaguchiOSM17} applies a simple mean pooling on the neighborhood structures of the emerging entities without distinguishing the edge types. 
To obtain a more effective neighborhood aggregator, LAN \cite{DBLP:conf/aaai/WangHLP19} employs two kinds of attention mechanisms, 
and the attention weights are estimated by either logic rules or a neural network.
These methods above still face the neighbor-sparseness problem when few neighbors of unseen entities are known. They also ignore the meaningful complex patterns in the KGs which only focuses on one-hop neighbors and fail to recognize the interactions between rules and embeddings.
In contrast, our method, VN network, proposes the concepts of the virtual neighbors to handle the sparseness problem, 
identify logic and symmetric path rules to capture more information, and optimize the KG embeddings and rule inference in an iterative manner.

\begin{figure*}
  \centering
  \includegraphics[width=6.5in]{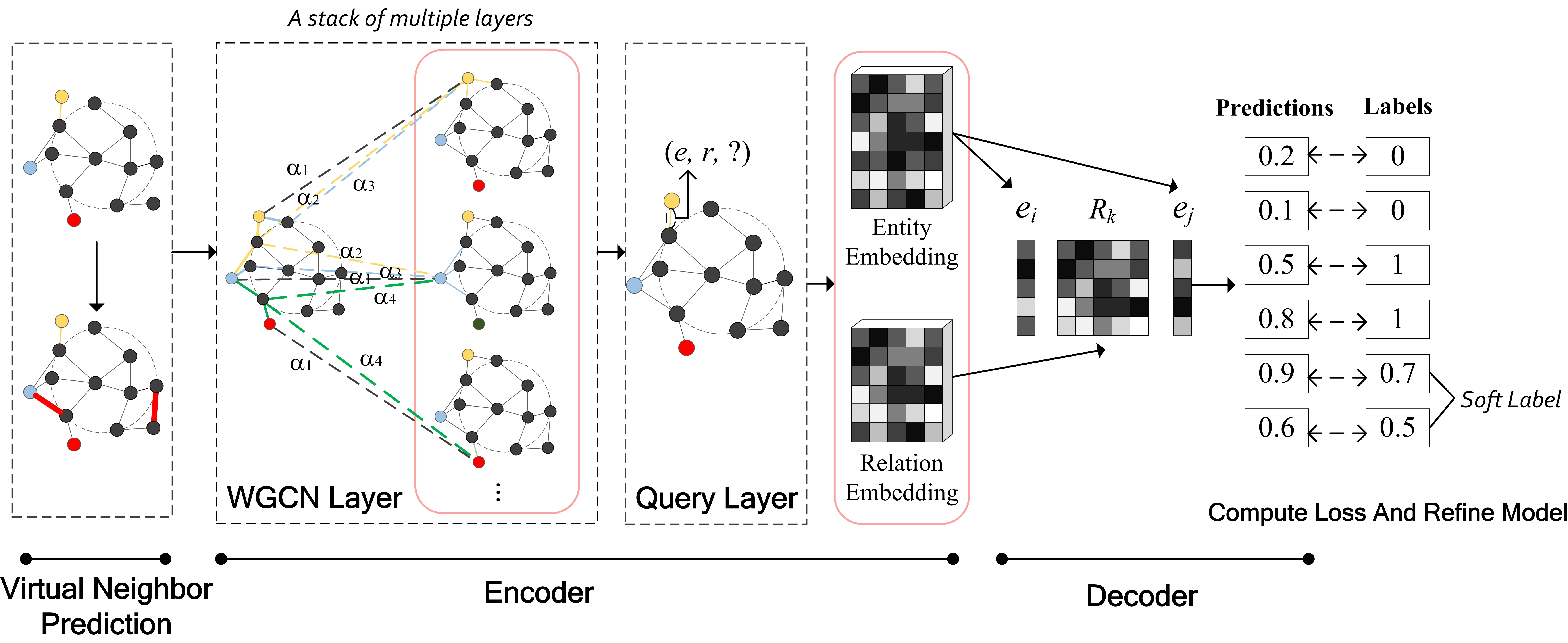}
  \caption{The framework of the VN network. VN network contains three components: the rule-based virtual neighbor prediction, the GNN-based encoder to capture structure information and embed entities, 
the decoder to calculate the probability of edges and then refine the model with soft labels.}
  \label{framework}
  \Description{The framework of the VN network.}
\end{figure*}
\section{Method}

This section introduces the proposed Virtual Neighbor network (VN network).
As Figure~\ref{framework} shows, the model has three main components: the rule-based virtual neighbor prediction,
a GNN-based encoder and an embedding-based decoder.
In the following sections, we describe the definitions and notations used in this paper firstly. 
And then we give an overview of the model architecture and detail the three components.

\subsection{Definitions}

This section introduces the definitions and notations for the knowledge graph (KG) under the setting of unseen entities and rules used in this paper.

\subsubsection{Knowledge Graph (KG)}
A knowledge graph can be considered as a multi-relational graph, consisting of a set of the \textbf{observed edges} (fact triples), i.e.,
$\mathcal{O} = \{x_{o}\}$,  where  $x_{o} = (e_{i}, r_{k}, e_{j})$. Each edge (triple) consists of two nodes (entities) $e_{i}, e_{j} \in \mathcal{E}_{o}$, 
and the edge (relation) $r_{k} \in \mathcal{R}$, where $\mathcal{E}_{o}$ and $\mathcal{R}$ are the entity and relation sets respectively. 
And for an entity $e$, we define its neighborhood in $\mathcal{O}$ as ${N}_{o}(e)$, where ${N}_{o}(e)=\{(e^{'},r)|(e^{'}, r, e) \in \mathcal{O} \lor (e, r, e^{'}) \in \mathcal{O}\}$.

In addition to the observed triples and entities, we collect the \textbf{newly emerging entities} and \textbf{auxiliary triples}.
Newly emerging entities are unseen during the training process but appear in the test set. 
And auxiliary triples are newly added facts for the original KG,
which connect the unseen entities with the original graph, to learn the embeddings for unseen entities during the aggregating process.
For example, as Figure~\ref{fig:examplesp} shows, $(London, isCityOf, England) \in \mathcal{O}$ is an observed triple, where $London, England \in \mathcal{E}_{\mathcal{O}}$ and $isCityOf \in \mathcal{R}$.
$(Alec, bornedIn, London) \in \mathcal{AUX}$ is an auxiliary triple, 
composed of one unseen entity $Alec \in \mathcal{E}_{u}$, one observed entity $London \in \mathcal{E}_{o}$, and their relation $bornedIn \in \mathcal{R}$,
where $\mathcal{AUX}$ is the auxiliary triple set, and $\mathcal{E}_{u}$ is the unseen entity set.

In the VN network, to address the sparseness problem of $\mathcal{AUX}$ as mentioned above, we introduce the concept of the \textbf{virtual neighbor},
which were obtained by using rules on $\mathcal{O}$ and $\mathcal{AUX}$ (detailed in \S \ref{sec:Virtual Neighbor Prediciton}).
To be more specific, a triple $(e_{i}, r_{k}, e_{j})$ is not in $\mathcal{O}$ and $\mathcal{AUX}$, where $e_{i} \in \mathcal{E}_{o}$, $e_{j} \in \mathcal{E}_{u}$ and $r_{k} \in \mathcal{R}$, 
we define the head entity $e_{i}$ as the \textbf{virtual neighbor} of the tail entity $e_{j}$ under relation $r_{k}$. 
We denote the set containing such kind of triples as $\mathcal{VN} = \{x_{\mathcal{VN}}\}$.
And as in Figure~\ref{fig:examplesp}, $(Alec, bornedIn, London)$ is newly added fact containing unseen entity "\emph{Alec}", 
so it belongs to $\mathcal{AUX}$.
$(Alec, starredIn, Star Wars)$ can be inferred by rules, and it is not in $\mathcal{O}$ and $\mathcal{AUX}$, 
so $(Star Wars$, $starredIn)$ is the virtual neighbor of "\emph{Alec}".

\subsubsection{Rule}
A set of rules with different confidence levels are denoted as $\mathcal{F}=\{f_{p},\lambda_{p}\}$,
where $f_{p}$ is the $p$-th logic rule defined over the given KG.
For example, $(i,r_{1},j) \Rightarrow (i,r_{2},j): i,j \in \mathcal{E}$ and $r_{1},r_{2} \in \mathcal{R}$, 
which indicates that any two entities connected by $r_{1}$ should also be connected by $r_{2}$.
The left-hand side of the implication $\Rightarrow$ is called the premise, and the right-hand side is conclusion. 
$\lambda_{p}$ represents the confidence level of rule $f_{p}$, where $\lambda_{p} \in$ [0,1].
The higher the confidence level of rule, the more likely it is.

\subsubsection{Symmetric Path}
Although logic rules can be mined through AMIE+, RLvLR or other useful rule mining methods,
which can help us find the missing facts, but they still have limitations.
As they mainly concentrate on one-hop or two-hop relations while there may be some long-distance dependencies helpful.
Moreover, due to the sparsity problem mentioned above, we may lack neighbors to mine sufficient rules or obtain the grounding rules.
Under such circumstance, \textbf{the symmetric paths} can help us find some entities that have similar contextual semantics, which contain two subpaths with the same relation order but the opposite direction.
As shown in Figure~\ref{fig:examplesp}, we start with the unseen entity "\emph{Alec}",
we can find two symmetric paths, namely, the blue one and the green one,
and the end node at the other end of the path is "\emph{Tom}" samely, but the two paths hint "\emph{Alec}" is an actor and a resident of England respectively.
Taking an example in meta-path learning \cite{sun2011pathsim}, 
A-P-A (representing co-authorship relationship) and A-P-V-P-A (representing two papers published by two authors in the same venue) as meta-paths for a bibliographic graph,
we can sample such node pairs and incorporate them in embeddings.
But above method is usually graph-specific and highly rely on knowledge from domain experts, which requires empirical results from previous works on the same graph.
Our proposed symmetric path do not have to learn such meta-path explicitly,
and semantically similar entities can be found to help enrich the neighbors for unseen entities.

\subsection{Model Architecture}

This section briefly introduces the model architecture with three components:
the rule-based virtual neighbor prediction, a GNN-based encoder, and an embedding-based decoder.
In the VN network, as Figure~\ref{framework} indicates, given a knowledge graph, 
we extract the logic and symmetric path rules to make the \textbf{virtual neighbor prediction}.
We predict a soft label $s(x_{vn})\in[0,1]$ to every unobserved triple containing virtual neighbors.
To do so, we solve a convex optimization problem constrained by rules and calculate the soft labels using the current KG embeddings.
As a result, the original KG becomes denser, which extremely facilitates embedding learning and predicting.

Then, the knowledge graph with virtual neighbors is inputted to the \textbf{GNN-based encoder}.
The key idea of the GNN-based encoder is to collect information from the neighbors and project nodes (entities) to the continuous spaces.
As mentioned in \cite{DBLP:conf/iclr/XuHLJ19}, modern GNNs follow the neighbor aggregation strategy, which is to update the representation of a node by aggregating representations of its neighbors repeatedly.
For a multi-relational knowledge graph, we can formulate the $l$ th layer of a GNN as:
\begin{equation}
    \label{eq:basic}
    \begin{split}
        a_{i}^{(l)} &= {\rm AGGREGATE}^{(l)}(h_{r,j}^{(l-1)}:(i, r, j) \in \mathcal{O}), \\
        h_{i}^{(l)} &= {\rm COMBINE}^{(l)}(h_{r_{0},i}^{(l-1)}, a_{i}^{(l)}),
    \end{split}
\end{equation}
where $a_{i}$ is the neighborhood aggregating information for the node $i$. $h_{r,j}^{(l)}$ denotes the message passing from the neighbor entity $j$ under relation $r$ at $l$ th layer,
$h_{r_{0}, i}^{(l)}$ denotes the self-connection message at $l$ th layer, 
and $r_{0}$ means the self-connection relation.
In this work, the encoder is composed of several \textbf{structure aware layers} and one \textbf{query aware layer}. 

Since knowledge base completion task aims to predict new facts when given an incomplete knowledge graph,
this task requires to determine how likely the unseen facts are true.  
To handle this task, the \textbf{decoder} should assign scores to the fact triples with the entity embedding $e_{i}$ from the GNN-based encoder, where $e_{i} = h_{i}^{L}: i\in \mathcal{E}_{o}$.
As the choice of the decoder is independent of the encoder, various methods can be used here.
In this work, we mainly adopt DistMult \cite{yang2014embedding} as our decoder,
and we also consider TransE \cite{DBLP:conf/nips/BordesUGWY13} and ComplEX \cite{DBLP:conf/icml/TrouillonWRGB16} decoders.

Finally, we optimize a global loss over observed facts and virtual neighbors to \textbf{refine embeddings}.
In that way, we can obtain unseen entities' embeddings fitting the ground truths while satisfying the extracted rules.
Note that, in the testing process, we only apply the rule-based virtual neighbor prediction and the GNN-based encoder on the auxiliary triple set $\mathcal{AUX}$,
and then use the obtained embeddings to predict missing facts about newly emerging entities without refining the soft labels since the testing process is not iterative.

\subsection{Virtual Neighbor Prediction}
\label{sec:Virtual Neighbor Prediciton}

In this section, we describe how to predict the soft label $s(x_{vn})$ for each triple $x_{vn}$ in $\mathcal{VN}$ inferred by the logic or symmetric path rules.

\subsubsection{Logic and Symmetric Path Rules}
For the logic rules obtained by the KG rule mining tool, 
we instantiate them with concrete entities to obtain the ground rules.
For example, given a rule ($i$, $playsFor$, $j$) $\Rightarrow$ ($i$, $isAffiliatedTo$, $j$), 
we can instantiate this rule as $(Saidu, playsFor,$ $Chiasso)$ $\Rightarrow$ $(Saidu, isAffiliatedTo, Chiasso)$.
As there could be a huge number of groundings and our goal is to reduce the sparsity problem for unseen entities,
we take as valid groundings only those where premise triples are observed in $\mathcal{O} \cup \mathcal{AUX}$ and conclusion triples that including the unseen entities in $\mathcal{E}_{u}$.
Moreover, we only use the rule whose confidence is not less than our threshold,
and the confidences of the rules are calculated directly by the AMIE+.

To get the symmetric path rule (SP rule) and its grounding, we do the following operations.
Firstly, starting from one unseen entity $e_i$, we search all the symmetric paths $SP(e_i)$ by random walks.
And if we find that the path is not symmetric, we terminate the search of this path early.
Then we consider the pairwise combinations of symmetric paths,
and the confidence level of them is calculated by the head coverage value \cite{DBLP:journals/vldb/GalarragaTHS15},
which mainly focus on the co-occurrence frequency of the premise and conclusion.
As shown in Table~\ref{Tab:sp}, the confidence of the $sp$ rule $sp_{i} \Rightarrow sp_{j}$ about "\emph{Bob}"  is 0.8, and we limit the number of $sp_i$ to no less than 5.

\begin{figure}[h]
  \centering
  \includegraphics[width=3.1in]{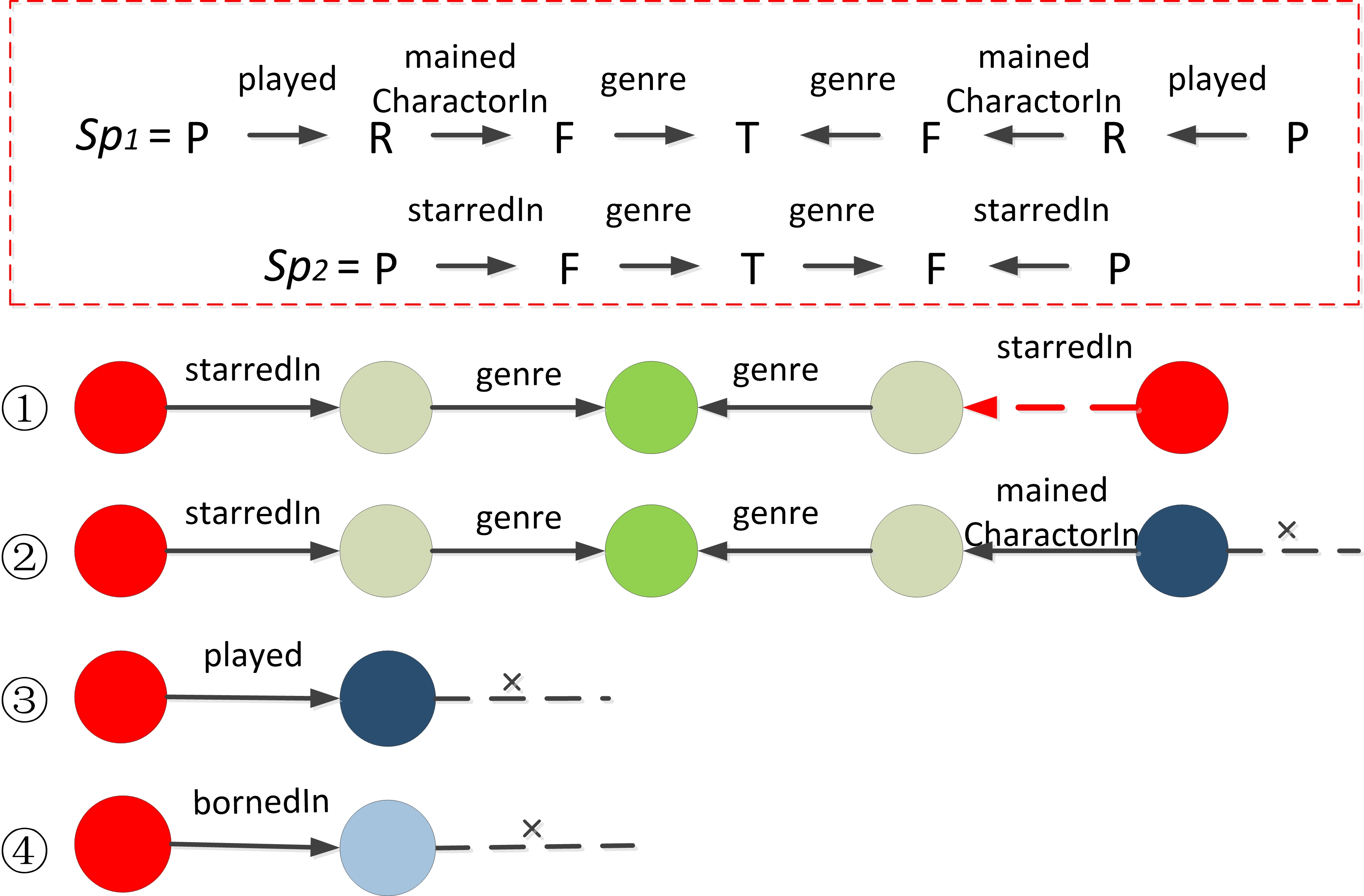}
  \caption{The example of the the symmetric path rule. P is person. R is role. F is film. And T is type.
Note that for symmetric paths, we only care about the type and direction of the edges, and the letters of the nodes in the graph are just to illustrate the underlying meaning of them.}
  \Description{The example of symmetric rule.}
  \label{sp2}
\end{figure}

\begin{table}
\caption{
An Example of KG Containing Two Types of Symmetric Path Starting from "\emph{Bob}"}
 \label{Tab:sp}    
    \begin{tabular}{c|c|c}  
    \toprule
    Node pair & $sp_i$ & $sp_j$\\
    \midrule
    (Bob, Zurich) & $\surd$ & $\surd$\\
    (Bob, Sam) & $\surd$ & $\times$\\
    (Bob, Rome) & $\surd$ & $\surd$\\
	(Bob, Amy) & $\surd$ & $\surd$\\
	(Bob, Adam) & $\surd$ & $\surd$\\ 
    \bottomrule
    \end{tabular}
\end{table}
For each symmetric path in the form of $sp_{i}$,
we search from the other end of it.
And if we can obtain the symmetric path in the form of $sp_{j}$ by connecting the last edges with the unseen entity $e_i$,
we get the grounding rule of $sp_{i} \Rightarrow sp_{j}$.
In Figure~\ref{sp2}, we add the edge "\emph{starredIn}" to obtain the $sp_{j}$ path which share the same ends of $sp_{i}$.
To model the dependency between the two end entities and for brevity,
we express the grounding rule of it as $(x_{first1})\land(x_{last1})\land({x_{last2}})\Rightarrow(x_{first2})$,
where $x_{first1}$ and $x_{last1}$ are the first triple and last triple in $sp_i$, $x_{last2}$ is the triple connect the other end of $sp_j$,
and $x_{first2}$ is the triple inferred as Figure~\ref{sp2} in $sp_j$.
In this way, we can focus on specific entities and capture more information from long-distance nodes, to assist with logic rules to enrich the neighbors for unseen entities.
We will demonstrate the effectiveness of SP rules through the experimental results.

\subsubsection{Soft Label Prediction}
To model rules, we adopt t-norm fuzzy logics \cite{DBLP:books/kl/Hajek98}. 
Following \cite{DBLP:conf/aaai/GuoWWWG18}, 
the logical conjunction ($\land$), disjunction($\lor$), and negation($\neg$) are defined as follow:
\begin{equation}
\label{eq:rules}
\begin{split}
    \pi(f_{1} \land f_{2}) &= \pi(f_{1}) \cdot \pi(f_{2}),\\
    \pi(f_{1} \lor f_{2}) &= \pi(f_{1}) + \pi(f_{2}) - \pi(f_{1}) \cdot \pi(f_{2}),\\
    \pi(\neg f_{1}) &= 1 - \pi(f_{1}),
\end{split}
\end{equation}
Here, $f_{1}$ and $f_{2}$ are two logical expressions, which can either be single triples or be constructed by combining triples with logical connectives.
$\pi(\cdot)$ assigns a truth value to each expression, indicating to what degree the logical expression is true.
For triples, $\pi(\cdot)$ is the score function (DistMult) used in the decoder.
For a conjunction of several triples,
the truth value can be calculated recursively through Eq.~\ref{eq:rules}. 
For example, for the rule $f_{i} \Rightarrow f_{j}$, the truth value can be computed as follows:
\begin{equation}
    \pi(f_{i} \Rightarrow f_{j}) = \pi(\neg f_{i} \lor f_{j}) = \pi(f_{i}) \cdot \pi(f_{j}) - \pi(f_{i}) + 1.
\end{equation}

To this end, our goal is to find a \textbf{soft label} $s(x_{vn}) \in S$ for each triple $x_{vn} \in \mathcal{VN}$,
using the current KG embeddings $\Theta$ and the ground rules $\mathcal{F}_{rule}$.
The optimal soft label $s(x_{vn})$ should keep close to truth value $\pi(x_{vn})$, while constrained by the ground rules.
To do so, we introduce the conditional truth value $\pi(f_{rule}|S)$ for the ground rule $f_{rule} \in \mathcal{F}_{rule}$. 
For example, for the ground rule $f_{rule} :(e_i,r_s,e_u) \Rightarrow (e_j,r_k,e_u)$, where premise $(e_i,r_s,e_u)$ is an known triple with unseen entity $e_u$,
and conclusion is an unknown triple $(e_j,r_k,e_u) \in \mathcal{VN}$ with unseen entity $e_u$.
Then, we can calculate $\pi(f_{rule}|S)$ as:
\begin{equation}
    \pi(f_{rule}|S) = I(e_i,r_s,e_u)\cdot s(e_j,r_k,e_u) - I(e_i,r_s,e_u) + 1,
\end{equation}
where $I(e_i,r_s,e_u)$ is a truth value defined by the score function with the current embeddings,
and $s(e_j,r_k,e_u)$ (can also be written as $s(x_{vn})$) is a soft label to be predicted for the triple containing the virtual neighbor.
To get the optimal soft labels, we introduce the slack variables $\xi_{f}$ for the rule $f$, and establish the following optimization problem:
\begin{equation}
\begin{split}
    \min_{S, \xi} &{\frac{1}{2}\sum_{x_{vn} \in \mathcal{VN}}{(s(x_{vn})-I(x_{vn}))^2}}+C\sum_{f \in F_{rule}}{\xi_{f}},\\
    s.t. &\lambda_{f}(1 - \pi(f|S)) \leq \xi_{f}, \xi_{f} \geq 0, 0\leq s(x_{vn})\leq 1,
\end{split}
\end{equation}
This kind of the optimization problem is convex \cite{DBLP:conf/aaai/GuoWWWG18}. 
Therefore, we can obtain the closed form solution:
\begin{equation}
\label{soft}
    s(x_{vn}) = \left[I(x_{vn}) + C\sum_{f \in F_{rule}}{\lambda_{f}\nabla_{s(x_{vn})}\pi(f|S)}\right]_{0}^{1}
\end{equation}
where $C$ is the constant penalty parameter, and $\lambda_{f}$ is the confidence value for the rule $f$, determined by the extracted algorithm.
$[\cdot]_{0}^{1} = \min(\max(x, 0), 1)$ is a truncation function.
Through using currently learned embeddings and rules to predict soft labels for virtual triples,
we consider the influence of embeddings on the rule-inference results rather than regarding them as necessarily true.

\subsection{Encoder and Decoder}
This section details the encoder and decoder in our framework.

\subsubsection{Structure and Query Aware Layers} 
As mentioned above, our encoder consists of several structure aware layers and one query aware layer.
In the first place, we adopt multiple GNN layers to map the connectivity structures of the KG into continuous spaces. 
To be specific, we use the weighted graph convolutional network (WGCN) \cite{DBLP:conf/aaai/ShangTHBHZ19} in the local aggregation process.
Each structure aware layer assigns the different attention weights for each relation type,
and the output embedding of the $l$ th layer for the entity $i$ can be formulated as follow:
\begin{equation}
\label{eq:WGCN}
\begin{split}
a_{i}^{(l)} &=  W^{(l)}(\sum_{(i,r,j) \in \mathcal{O}}{\alpha_{r}^{(l)}h^{(l-1)}_{j}}),\\
h_{i}^{(l)} &=  \tanh(a_{i}^{(l)} + h_{i}^{(l-1)}W^{(l)}),
\end{split}
\end{equation}
where $\alpha_{r}$ is the attention weight for relation $r$ connected with the head entity $i$ and tail entity $j$.
$h_{i}^{(l)} \in R^{d^{(l)}}$ is the embedding of entity $i$ at the $l$ th layer. 
$W^{(l)} \in R^{d^{(l-1)} \times d^{(l)}}$ is the connection matrix for the $l$ th layer, transforming $h^{(l-1)}_{i}$ to $h^{(l)}_{i}$.
We randomly initialize the input entity embedding $h^{(0)}_{i}$, 
and stack multiple structure aware layers before the query aware layer. 

Besides the common structure information in the KG, 
given the query relation (relation of the inputted triple), an ideal aggregator is able to focus on the relevant facts in the neighborhood \cite{DBLP:conf/aaai/WangHLP19}.
To exploit the query-relevant information, we construct a query aware layer based on the neural network mechanism \cite{DBLP:conf/aaai/WangHLP19}. 
Specifically, given a query relation $q\in \mathcal{R}$, the importance of the neighbor $j$ to entity $i$ is calculated as follow:
\begin{equation}
\label{eq:NN attention}
\alpha^{\rm NN}_{j|i, q} = {\rm softmax}(\beta_{j|i,q}) = \frac{{\rm exp}(\beta_{j|i,q})}{\sum_{(i,r,j')\in \mathcal{O}}{{\rm exp}(\beta_{j'|i,q})}},
\end{equation}
where the unnormalized attention weight $\beta_{j|i,q}$ can be computed by the following neural network:
\begin{equation}
    \beta_{j|i, q} = {\rm LeakyReLU}(u\cdot[W_{e}h_{i};W_{q}z_{q};W_{e}h_{j}]),
\end{equation}
where $u\in R^{3d}$, $W_{e}$ and $W_{q} \in R^{d\times d}$ are the global attention parameters, 
while $z_{q}\in R^{d}$ is a relation-specific attention parameter.
Then, given the query relation $q$, the whole query aware layer can be defined as:
\begin{equation}
    \label{eq:query aware layer}
    h^{O}_{i}= \sum_{(i,r,j)\in \mathcal{O}}{\alpha^{\rm NN}_{j|i, q} \cdot h^{I}_{j}},
\end{equation}
where $h^{I}_{j} = h^{(L)}_{j}$ is the embedding of the entity $j$ from the last structure aware layer.
$h^{O}_{i} = e_{i}$ is the output embedding of the entity $i$ to the decoder.
When obtaining embeddings for unseen entities, we apply the encoder on auxiliary and virtual triples by simply replacing all the observed triples $(i, r, j)\in \mathcal{O}$ in Eq. \ref{eq:WGCN}, \ref{eq:NN attention} and \ref{eq:query aware layer} with the triples $(i', r', j')\in \mathcal{AUX}\cup\mathcal{VN}$.

\subsubsection{DistMult Decoder} 
Then we use DistMult as our decoder to assign scores for the fact triples with the entity embedding $h^{O}_{i}$ output from the GNN-based encoder:
\begin{equation}
    \label{eq:DistMult}
    I(e_{i}, r_{k}, e_{j}) = e_{i}^{T}R_{k}e_{j},
\end{equation}
where $I(\cdot)$ is the score function. $e_{i}, e_{j} \in \Theta$ are the entity embeddings, and $R_{k}$ is a diagonal matrix for the relation $r_{k} \in \mathcal{R}$.

\begin{algorithm}[htb]
\caption{Procedure of VN Network}
\label{alg:process}
	\begin{algorithmic}[1]
	\REQUIRE {\ \\
	The positive triples and negative triples $\mathcal{L}=\{(x_{l}, y_{l})\}$;\\
	The triples inferred by rules that containing virtual neighbors $\mathcal{VN} = \{x_{\mathcal{VN}}\}$;\\
	The trained encoder and decoder on the original KG;\\
	The iteration number $N$;}\\
	\REPEAT \FOR {each mini-batch $\mathcal{L}_b$, $\mathcal{VN}_b$}
	\STATE Compute the soft labels for $\mathcal{VN}_b$ using Eq. \ref{soft};\ 
	\STATE Input $\mathcal{L}_b$, $\mathcal{VN}_b$ into encoder;
	\STATE Get the scores for $\mathcal{L}_b$ and  $\mathcal{VN}_b$ from decoder;
	\STATE Minimize the global loss using Eq. \ref{eq:loss};
	\STATE $n \gets n + 1$
	\ENDFOR
	\UNTIL{$n < N$}
	\ENSURE {$\Theta^N$\\}           
	\end{algorithmic}
\end{algorithm}

\subsection{Training Objective}

Algorithm 1 summarizes the learning procedure of our method.
We randomly corrupt the head or tail entity of a positive triple to form negative triples.
We are given a hard labeled set $\mathcal{L}=\{(x_{l}, y_{l})\}$, 
each positive or negative triple in it has a hard label $y_{l} \in \{0, 1\}$.
We also collect the set of virtual triples with soft labels, i.e., $\mathcal{VN} = \{x_{\mathcal{VN}}\}$.
Our goal is to learn the optimal KG embeddings $\Theta^N$ with both hard labeled and soft labeled triples.
To do so, we establish a loss function over $\mathcal{L}$ and $\mathcal{VN}$ as follow:
\begin{equation}
\label{eq:loss}
\min_{\Theta}{\frac{1}{|\mathcal{L}|}\sum_{\mathcal{L}}{l(I(x_{l}), y_{l})} + \frac{1}{|\mathcal{VN}|}\sum_{\mathcal{VN}}{l(I(x_{vn}), s(x_{vn}))}},
\end{equation}

where we adopt the cross entropy $l(x,y) = -y\log{x} - (1-y)\log{(1-x)}$.
$I(\cdot)$ is the score function defined in Eq.\ref{eq:DistMult}.
We use the ADAM algorithm \cite{DBLP:journals/corr/KingmaB14} to minimize the global loss function.
In this way, the resultant embeddings of unseen entities fit the newly emerging facts while constrained by rules.

\begin{table}
\caption{
The Statistics of Three Datasets}
 \label{Tab:threedataset}
    \resizebox{\linewidth}{!}{
    \begin{tabular}{c|cc|ccc}  
    \toprule
    Dataset & Entities & Relations & Training & Validation & Test\\
    \midrule
    WordNet11 & 38,696 & 11 & 112,581 & 5,218 & 21,088\\
    FB15k & 14,951 & 1,345 & 483,142 & 50,000 & 59,071\\
    YAGO37 & 123,189 & 37 & 989,132 & 50,000 & 50,000\\ 
    \bottomrule
    \end{tabular}
    }
\end{table}

\section{Experiments}

In this section, we evaluate our proposed framework, VN network, in two KBC tasks: triple classification and link prediction.
We mainly compare our model with three baselines, LSTM and LAN in \cite{DBLP:conf/aaai/WangHLP19}, and MEAN in \cite{DBLP:conf/ijcai/HamaguchiOSM17}.
And we mainly evaluated our method from the following perspectives,
1)whether our model can learn better embeddings for unseen entities than the above methods with the neighbor sparsity problem,
2)whether each component of our framework is useful for the learning of embeddings.

\begin{table*}
  \caption{
Statistics of Processed FB15K and YAGO37 Dataset}
  \label{Tab:FB15k Datasets}
	\resizebox{\textwidth}{!}{
    \begin{tabular}{c|ccc|cc|cc|ccc|cc|cc}  
    \toprule
			&\multicolumn{7}{c|}{FB15K}&\multicolumn{7}{c}{YAGO37}\\  
	Dataset & $\mathcal{O}$ & $\mathcal{AUX}$ & $\mathcal{E}_{u}$ & Valid & Test & $avg-\mathcal{R}_{before}$ & $avg-\mathcal{R}_{after}$  & $\mathcal{O}$ & $\mathcal{AUX}$ & $\mathcal{E}_{u}$ & Valid & Test & $avg-\mathcal{R}_{before}$ & $avg-\mathcal{R}_{after}$\\  
    \midrule
    Subject-5 & 188,238 & 235,746 & 1,465 & 19,454 & 1,834 & 85.54 & 125.89 & 929,037 & 59,453 & 2,408 & 49,972 & 2,458 & 23.79 & 35.04\\
    Object-5 & 170,672 & 254,454 & 1,344 & 17,712 & 1,896 & 88.66 & 129.51 & 568,900 & 269,988 & 1,774 & 49,933 & 2,484 & 140.22 & 146.71\\ 
    \midrule
    Subject-10 & 108,854 & 249,798 & 2,102 & 11,339 & 2,811 & 52.33 & 73.31  & 899,110 & 87,868 & 4,707 & 49,861 & 4,925 & 17.26 & 27.92\\
    Object-10 & 99,783 & 261,341 & 1,947 & 10,190 & 2,987 & 53.63 & 74.58 & 441,916 & 382,082 & 3,188 & 49,815 & 4,926 & 92.58 & 99.61\\ 
    \midrule
    Subject-15 & 71,407 & 228,484 & 2,358 & 7,310 & 3,250 & 37.14 & 50.36 & 848,034 & 137,255 & 6,846 & 49,752 & 7,299 & 17.43 & 22.49\\
    Object-15 & 67,651 & 243,316 & 2,228 & 6,878 & 3,703 & 39.00 & 52.65 & 380,053 & 440,759 & 4,499 & 49,699 & 7,379 & 67.56 & 75.31\\
    \midrule
    Subject-20 & 49,456 & 205,242 & 2,571 & 5,048 & 3,586 & 29.19 & 38.37 & 812,019 & 170,536 & 8,838 & 49,614 & 9,601 & 15.93 & 26.36 \\
    Object-20 & 46,982 & 222,200 & 2,388 & 4,843 & 4,132 & 30.92 & 40.59  & 332,204 & 488,332 & 5,884 & 49,556 & 9,745 & 55.25 & 63.71\\
    \midrule
    Subject-25 & 37,986 & 179,656 & 2,704 & 3,908 & 3,889 & 22.53 & 29.12 & 781,209 & 200,073 & 10,897 & 49,526 & 12,048 & 15.76 & 26.18\\
    Object-25 & 34,126 & 195,627 & 2,470 & 3,498 & 4,283 & 24.92 & 32.01 & 297,655 & 523,251 & 7,234 & 49,452 & 12,193 & 45.69 & 51.37\\ 
    \bottomrule
    \end{tabular}}
\end{table*}

Since the link prediction is a common task for knowledge graph completion which considers the ranks of all entities,
we do further analysis on the link prediction task.
\subsection{Datasets}

We evaluate VN network on three datasets: WordNet11 \cite{socher2013reasoning}, FB15k \cite{DBLP:conf/nips/BordesUGWY13} and YAGO37 \cite{DBLP:conf/aaai/GuoWWWG18}.
WordNet11 is a subset of WordNet, which is a database of lexical relations between words.
FB15K is a subset of the multi-relational knowledge base Freebase. 
And YAGO37 is extracted from the core facts of YAGO3 where entities appearing less than 10 times are discarded.
Table~\ref{Tab:threedataset} summarizes the detail statistics of the above datasets.

For the triple classification task, we directly use the datasets released in \cite{DBLP:conf/ijcai/HamaguchiOSM17} based on WordNet11, including Subject-\{1000, 3000, 5000\}, Object-\{1000, 3000, 5000\} and Both-\{1000, 3000, 5000\}.
For the link prediction task, the published data sets of FB15K differ from what is written in \cite{DBLP:conf/aaai/WangHLP19}, and there is no public dataset available on YAGO37 under the unseen entities setting.
So we use the datasets on FB15K that they publish,
and construct the datasets on YAGO37 by following the similar protocol mentioned in \cite{DBLP:conf/aaai/WangHLP19}, including Subject-\{5, 10, 15, 20, 25\} and Object-\{5, 10, 15, 20, 25\}.
The basic idea of the dataset construction is to randomly sample triples from test set, and select entities from these triples as candidate unseen entities.
Then using these entities to split the training set.
The specific process is as follows:

 \textbf{Sampling unseen entities}. 
 Firstly, $R = {5\%,10\%,15\%,20\%,25\%}$ triples are randomly sampled from the original FB15K(YAGO37) test set as candidate test set.
 As for WordNet11, $N = \{1000,3000,5000\}$ testing triples are sampled.
 And Subject is the strategy that used to construct the candidate unseen entities sets $\mathcal{E}_{u}$, where only entities appearing as the head entities in the candidate test set are added to $\mathcal{E}_{u}$.
 The same goes for Both and Tail, where tail entities or both the head and tail entities are added to $\mathcal{E}_{u}$.
 Then entities in $\mathcal{E}_{u}$ that do not have any neighbors in the original training set are filtered.
 For a triple  $(e_{i}, r_{k}, e_{j})$ in the candidate test set, if $e_{i} \notin \mathcal{E}_{u} \land e_{j} \notin \mathcal{E}_{u}$, it is removed from the candidate test set.
 The new test sets $\mathcal{T}$ are obtained after filtering.
 
 \textbf{Filtering and splitting data sets}. 
 The second step is to ensure that unseen entities would not appear in the final training set or validation set. 
 The original training set are split into two data sets, the new training set $\mathcal{O}$ and auxiliary set $\mathcal{AUX}$. 
 For a triple $(e_{i}, r_{k}, e_{j})$ in original training set, if $e_{i} \in \mathcal{E}_{o} \land e_{j} \in \mathcal{E}_{o}$, it is added to the new training set $\mathcal{O}$.
 If $e_{i} \in \mathcal{E}_{o} \land e_{j} \in \mathcal{E}_{u}$ or $e_{i} \in \mathcal{E}_{u} \land e_{j} \in \mathcal{E}_{o}$, it is added to the auxiliary set $\mathcal{AUX}$, which serves as existing neighbors for unseen entities in the aggregating process.
For the new validation set, we keep only triples that have no unseen entities.

We employ AMIE+ \cite{DBLP:journals/vldb/GalarragaTHS15} to extract the logic rules on all the datasets.
We only keep the logic rules with the length not longer than 2 and the confidence not less than 0.8.
Besides, through the use of random walks and drool rule engine tool \cite{browne2009jboss},
we extract symmetric paths with length 2, 4 and 6, and also keep the symmetric path rules with the confidence not less than 0.8.
Then we enrich the neighbors for unseen entities with these rules.
We omit the virtual neighbors that overlap with the existing neighbors, and take the one with the highest confidence when the virtual neighbors conflict with themselves.
The detail statistics of WordNet11 can be found in \cite{DBLP:conf/ijcai/HamaguchiOSM17}.
And Table~\ref{Tab:FB15k Datasets} summarize the detail statistics of FB15K and YAGO37.
Since most of the tail entities of YAGO37 are locations or organizations which have many connected edges, so the subject-$R$ and object-$R$ are not balanced as FB15K.
We can see that the average ratio of neighbors to predicted facts of unseen entities has increased by at least five after the supplement, 
which helps to solve the sparsity problem.
\begin{table}
\caption{
Evaluation Accuracy on Triple Classfication(\%)}
 \label{Tab:tripleclassification}
    \resizebox{\linewidth}{!}{
    \begin{tabular}{c|ccc|ccc|ccc}  
    \toprule
          &\multicolumn{3}{c|}{Subject}&\multicolumn{3}{c|}{Object}&\multicolumn{3}{c}{Both}\\
    Model &1000 &3000 &5000 &1000 & 3000& 5000 & 1000 & 3000 & 5000\\        
    \midrule
    MEAN &87.3 &84.3 & 83.3 &84.0 &75.2 & 69.2 &83.0 &73.3 &68.2\\
    LSTM &87.0 &83.5 & 81.8 &82.9 &71.4 &63.1 &78.5 &71.6 &65.8\\
    LAN & 88.8 &85.2 & 84.2 &84.7 &78.8 &74.3 &83.3 &76.9 &70.6\\ 
    \midrule
    VN network & \textbf{89.1} & \textbf{85.9} &\textbf{85.4} &\textbf{85.5} &\textbf{80.6} &\textbf{76.8} &\textbf{84.1} &\textbf{78.5} &\textbf{73.1}\\
    \bottomrule
    \end{tabular}
    }
\end{table}

\subsection{Triple Classification Task}

This task is to classify every test triple as true or false. 
To tackle this task, we need to set a threshold $\delta_{r}$ for each relation $r$.
The triple $x=(e_{i},r,e_{j})$ is predicted to be true when $I(x)\geq \delta_{r}$, otherwise the triple is false.
We determine the optimal $\delta_{r}$ by maximizing classification accuracy on the validation set.

\begin{table*}
  \caption{
The Link Prediction Results}
  \label{Tab:Linkprediction}
	\resizebox{\textwidth}{!}{
    \begin{tabular}{c|ccccc|ccccc|ccccc|ccccc}
    \toprule
		&\multicolumn{10}{c|}{FB15K}&\multicolumn{10}{c}{YAGO37}\\
	\cline{6-7} \cline{16-17}
        &\multicolumn{5}{c|}{Subject-10}&\multicolumn{5}{c|}{Object-10}  &\multicolumn{5}{c|}{Subject-10}&\multicolumn{5}{c}{Object-10}\\
	\cline{3-5} \cline{8-10} \cline{13-15} \cline{18-20}
    Model &MR &MRR &Hits@10 &Hits@3 & Hits@1&MR &MRR &Hits@10 &Hits@3 & Hits@1  &MR &MRR &Hits@10 &Hits@3 & Hits@1&MR &MRR &Hits@10 &Hits@3 & Hits@1\\
    \midrule
    MEAN &293 &31.0 &48.0 &34.8 &22.2 &353 &25.1 &41.0 &28.0 &17.1  &2393 &21.5 &42.0 &24.2 &17.8 &4763 &17.8 &35.2 &17.5 &12.1\\
    LSTM &353 &25.4 &42.9 &29.6 &16.2 &504 &21.9 &37.3 &24.6 &14.3  &3148 &19.4 &37.9 &20.3 &15.9 &5031 &14.2 &30.9 &16.1 &11.8\\
    LAN & 263 &39.4 &56.6 &44.6 &30.2 &461 &31.4 &48.2 &35.7 &22.7  & 1929 &24.7 &45.4 &26.2 &19.4 &4372 &19.7 &36.2 &19.3 &13.2\\ 
    \midrule
    VN network & \textbf{175} & \textbf{46.3} &\textbf{70.1} &\textbf{52.6} &\textbf{34.5} &\textbf{212} &\textbf{42.3} &\textbf{62.7} &\textbf{44.6} & \textbf{28.2}  & \textbf{1757} & \textbf{46.5} &\textbf{66.8} &\textbf{53.8} &\textbf{35.7} &\textbf{3145} &\textbf{27.4} &\textbf{50.1} &\textbf{36.4} & \textbf{19.5}\\
    \bottomrule
    \end{tabular}}
\end{table*}
\subsubsection{Experimental Setup}
We randomly sample 64 negative triples for each triple in $\mathcal{O} \cup \mathcal{AUX}$. 
For all the datasets, we create 100 mini-batches on each dataset, and we conduct a grid search to find hyperparameters that maximize accuracy on the validation sets in at most 100 iterations.
The embedding dimensions for the encoder and decoder are all set to 200. 
The learning rate is 0.002.
The dropout rate during training is 0.3.
The regularization penalty coefficient on KG embeddings is 0.001.
The constant penalty $C$ is 0.01.
We employ three structure aware layers and one query aware layer in the encoder, and DistMult in the decoder.
Our models are implemented by PyTorch and run on NVIDIA TITAN RTX Graphics Processing Units. 

\subsubsection{Results}
The detailed results are shown in Table~\ref{Tab:tripleclassification}. 
We can see that VN network achieves the best performance over all the datasets, 
and most of the improvements over other baselines are significant.
The results demonstrate that embedding newly emerging entities with virtual neighbors 
indeedly improves the quality of KG embeddings. Note that the improvements on Subject-\{1000,3000,5000\}
are not considerable as the other two groups. 
It may be because the Subject-$R$ datasets are much easier to classify than the other two groups, the performances are more difficult to improve.
Interestingly, with the number of unseen entities increase, the improvement is more obvious,
which proves that our method is more effective in dealing with more unseen entities.

\subsection{Link Prediction Task}

Link prediction task aims at completing every test triple with subject or object missing.
For example, in Subject-$R$(Object-$R$) data sets, we firstly hide the object(subject) of each testing triple to produce a missing part.
Then we replace the missing part with all entities to construct candidate triples. 
We compute the scoring function defined in Eq.~\ref{eq:DistMult} for all candidate triples, and rank them in descending order. 
Finally, we evaluate whether the ground-truth entities are ranked ahead of other entities.
We use three widely used metrics: \emph{mean rank} (MR), \emph{mean reciprocal rank} (MRR), \emph{Hits at n} (Hits@n).
And for this ranking process, we remove corrupted triples which is called "filtered” setting. 
We report filtered MRR and Hits at 1, 3 and 10.

\subsubsection{Experimental Setup}
In this task, we finetune the embedding dimension $d$ in $\{100, 150, 200, 250, 300\}$, the dropout rate $\alpha$ in $\{0.1$, $0.15$, $0.2$, $0.25$, $0.3$, $0.4\}$, 
the learning rate $\beta$ in $\{0.001$, $0.002$, $0.005$, $0.01$, $0.1\}$ with a grid search.
The optimal configurations are $d$ = 100, $\alpha$ = 0.25 and $\beta$ = 0.002 for FB15K.
And for YAGO37, the optimal configurations are $d$ = 200, $\alpha$ = 0.2 and $\beta$ = 0.002.
Other hyperparameters are set the same as those in the triple classification task.

\subsubsection{Results}
As Table~\ref{Tab:Linkprediction} shows, VN network outperforms all the baselines.
Compared to the best performing baseline LAN,
our model achieves an improvement of 13.5\% in HITS@10 on FB15K, and an improvement of 21.4\% on YAGO37 under the setting of Subject.
And our model achieves an improvement of 14.5\% in HITS@10 on FB15K, and an improvement oft 13.9\% on YAGO37 under the setting of Object. 
The link prediction results demonstrate the superiority of our proposed framework again.
To explore whether each component is useful in our VN network and generalizes to other configurations, we will do further analysis on the link prediction task in the following.

\begin{table}
  \caption{
Effectiveness of Each Component on Subject-10}
  \label{Tab:Effectiveness}
    \begin{tabular}{c|ccccc}  
    \toprule
    Model &MR &MRR &Hits@10 &Hits@3& Hits@1\\        
    \midrule
    Structure aware &291 &34.2 & 56.7 &40.3 &24.8\\
    Hard rules &237 &40.6 &64.2 &46.5 &30.3\\
    Logic rules &218 &43.9 &66.2 &50.6 &32.3 \\
    Logic and SP rules &185 &45.9 &69.8 &51.9 &33.6 \\
    \midrule
    VN network & \textbf{175} & \textbf{46.3} &\textbf{70.1} &\textbf{52.6} &\textbf{34.5} \\
    \bottomrule
    \end{tabular}
\end{table}
\textbf{Ablation Study:}
To investigate the effectiveness of the components in our framework.
We conduct the link prediction task on several variants of our method, as Table~\ref{Tab:Effectiveness} describes.
First, we only employ structure aware layers, which weighs the different types of relations differently.
Then, without the soft label prediction process, we directly inject the hard rules which always hold with no exception.
Next, to ensure the necessity of the soft label prediction, we adopt the soft logic rules setting.
Besides, we add the symmetric path rules for the logic and SP rules setting, to ensure whether using long-distance dependencies between entities can be helpful.
Finally, we evaluate our method with the query aware layer.

As Table~\ref{Tab:Effectiveness} shows, even if we only use the structure aware layers,
our method still outperforms the simple MEAN aggregator in most cases, which indicates that the structure aware layers are able to effectively encode the connectivity structure information.

Meanwhile, We can see two significant improvements,
that is due to the use of hard labeled neighbors inferred by hard rules and iteratively soft labeled virtual neighbors.
The first significant improvement strongly demonstrates the sparsity problem of unseen entities, and we can learn more expressive embeddings for the unseen entities by enriching the neighbors.
The second significant improvement shows the importance of adding soft labeled virtual neighbors in an iterative manner.
Because when we add virtual neighbors for unseen entities, we also introduce noise.
Therefore, through considering the interactions between KG embeddings and rules,
we can obtain embeddings that are more accurate and adapt to the rules.
\begin{table}[h]
  \caption{
Different Decoders on Subject-10}
  \label{Tab:Decoder}
  \resizebox{\linewidth}{!}{
    \begin{tabular}{c|c|cccc}  
   \toprule
    Encoder & Decoder &MRR &Hits@10 &Hits@3 & Hits@1\\        
    \midrule
    MEAN & ComplEx & 28.6 & 44.7 & 32.2 & 20.4 \\
    MEAN & TransE  & 31.0 & 48.0 & 34.8 & 22.2\\
    MEAN & DisMult & 29.7 & 45.8 & 33.5 & 21.2\\
    \midrule
    LAN & ComplEx & 37.1 & 53.1 & 42.2 & 28.7 \\
    LAN & TransE & 39.4 & 56.6 & 44.6 & 30.2\\
    LAN & DisMult & 37.8 & 53.4 & 43.2 & 29.3 \\
    \midrule
    VN & ComplEx & 39.6 & 60.1 & 45.4 & 29.7\\
    VN & TransE & 43.5 & 66.7 & 49.4 & 31.6\\
    VN & DisMult & 46.3 & 70.1 & 52.6 & 34.5\\
    \bottomrule
    \end{tabular}}
\end{table}

Although the use of symmetric path rules does not improve as the two components above,
the improvement proves that the representational power of the original embeddings can be improved by introducing the long-distance dependencies between entities.
Our proposed framework achieves the best result by the use of the query aware layer,
and the main improvements are Hits@1 and Hits@3,
because it can exploit the query-relevant information which helps to focus on more relevant facts in the neighborhood.
\begin{figure}[h]
    \centering
    \subfigure[FB15K-Subject-R.]{
        \centering
    \label{fig:subject-R}
    \includegraphics[width=1.4in]{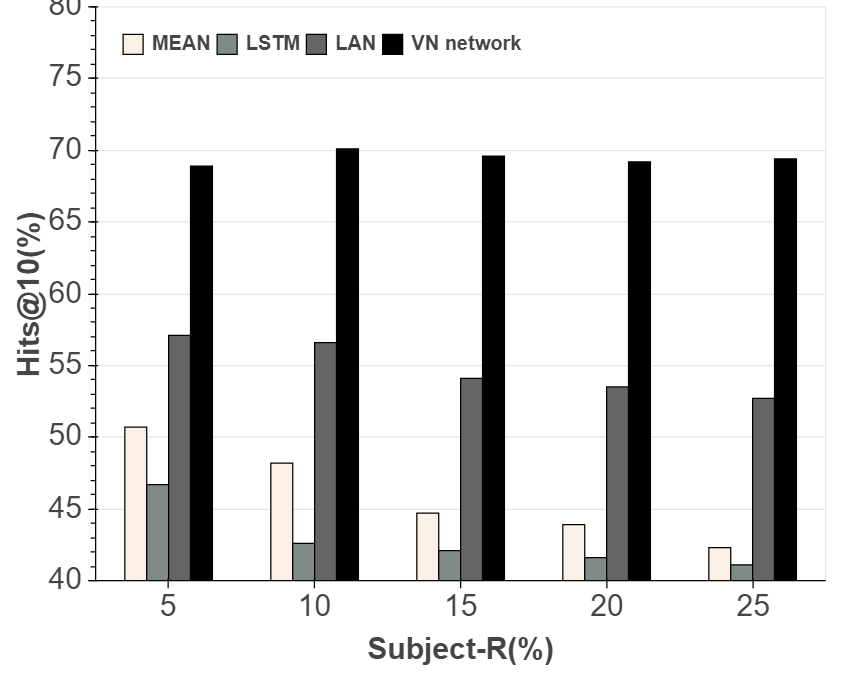}
    }    
    \subfigure[FB15K-Object-R.]{
        \centering
    \label{fig:object-R}
    \includegraphics[width=1.4in]{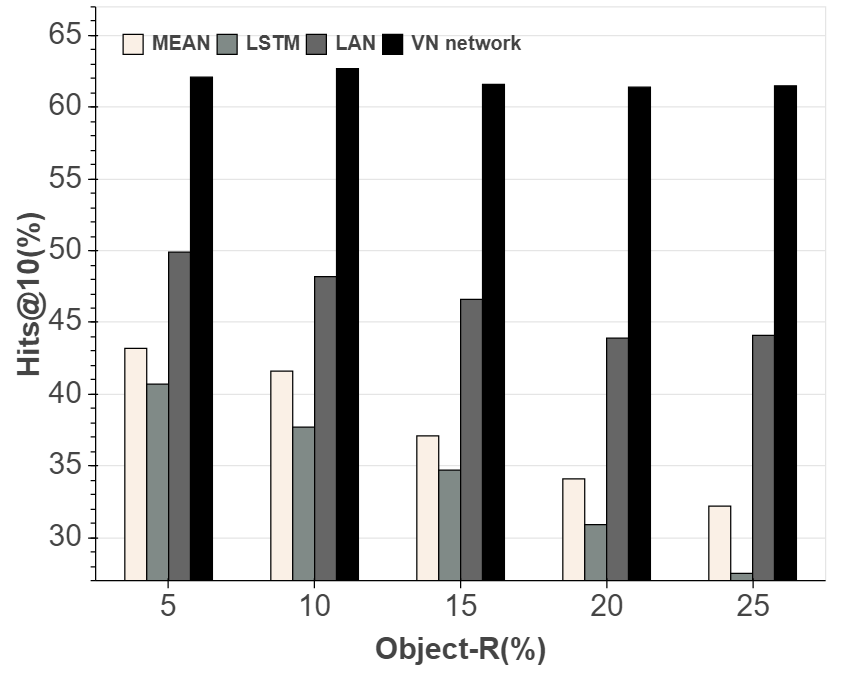}
    }
	\subfigure[YAGO37-Subject-R.]{
        \centering
    \label{fig:yagosubject-R}
    \includegraphics[width=1.4in]{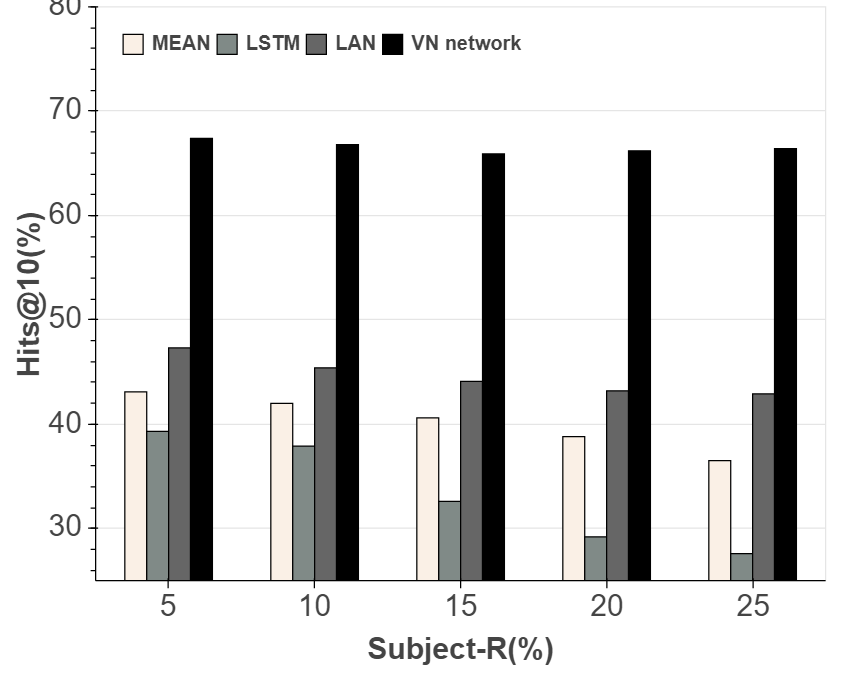}
    }    
    \subfigure[YAGO37-Object-R.]{
        \centering
    \label{fig:yagoobject-R}
    \includegraphics[width=1.4in]{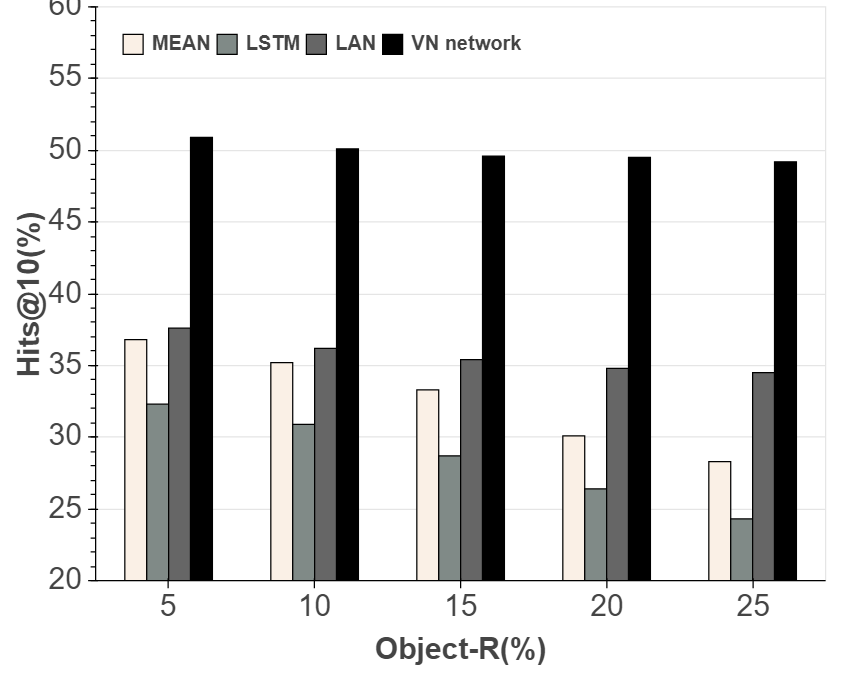}
    }
    \caption{Link prediction results on Subject/Object-R.}
    \label{fig:subject/object} 
\end{figure}

\textbf{Impact of Other Decoders:}
To find out the influence of the different decoders, we consider three typical embedding methods here: DistMult, TransE and ComplEx.
The LSTM is still inferior to MEAN as described in \cite{DBLP:conf/aaai/WangHLP19}, so we also omit the results of LSTM.
The results are reported in Table~\ref{Tab:Decoder}.
We can see that our method outperforms MEAN consistently by a large margin on all the evaluation metrics.
And as for the LAN, VN network performs better on most metrics.
VN network with the ComplEx decoder results in the worst performance, due to the high parameter complexity.
In contrast, VN network with the DistMult or TransE decoder can achieve the state-of-the-art results, 
while the DistMult leads to the best result.
The experiment results show that the superiority of our model to the baselines can generalize to other scoring functions and learn more expressive embeddings for unseen entities.

\textbf{Impact of the Percentage of Unseen Entities:}
In order to investigate the impact of the percentage of unseen entities, we conduct experiment under the setting of the Subject-$R$ and Object-$R$ on the FB15K and YAGO37 datasets.
And it seems reasonable that with the ratio of the unseen entities over the training entities increases (namely the observed knowledge graph becomes sparser), the accuracy would decline.

As Figure~\ref{fig:subject/object} shows, our method and LAN are much better than MEAN and LSTM.
And with the percentage of the unseen entities increasing and the KG becoming spaser,
VN network still achieves better results than other baselines on all data sets.
We observe that the increasing proportion of unseen entities certainly has a negative impact on all models because of the sparsity problem as mentioned above,
especially MEAN and LSTM can not learn effective embeddings as LAN and VN network when the proportion of unseen entities is high.
But the improvements of our model are relatively stable.
And on the YAGO37 data sets, there only drops of less than 2\% from 5 percent to 25 percent in our model,
while LAN drops more obvious.
We can conclude that when unseen entities appear,
with the iterative guidance from virtual neighbors, 
our framework can reduce the sparsity problem and accurately predict the missing facts of the unseen entities to learn more expressive embeddings.

\section{Conclusions}

In this paper, we discuss the problems of the KG embedding task under the setting of unseen entities.
And we propose a novel framework, VN network, to address the unseen entity problems.
To handle the sparsity problem, we introduce the short logic rules and symmetric path rules to capture more information and enrich the neighbors for unseen entities.
We also use three structure aware layers and one query aware layer,
which can adapt the amount of information from neighbors used in local aggregation and concentrate on more relevant information, leading to more accurate embeddings of unseen entities. 
And through introducing the concept of the virtual neighbor and employ a rule-based prediction algorithm to assign soft labels using the current KG embeddings,
we consider the interactions between the rule predictions and KG embedding learning rather than making a one-time injection of logic rules.
Experimental results show that VN network achieves improvements over state-of-the-art baselines.

\begin{acks}
This work was supported by the National Key R\&D Program with No.2016QY03D0503,2016YFB081304, and Strategic Priority Research Program of Chinese Academy of Sciences, Grant No.XDC02040400.
\end{acks}

\bibliographystyle{ACM-Reference-Format}
\bibliography{sample-base}


\begin{thebibliography}{35}


\ifx \showCODEN    \undefined \def \showCODEN     #1{\unskip}     \fi
\ifx \showDOI      \undefined \def \showDOI       #1{#1}\fi
\ifx \showISBNx    \undefined \def \showISBNx     #1{\unskip}     \fi
\ifx \showISBNxiii \undefined \def \showISBNxiii  #1{\unskip}     \fi
\ifx \showISSN     \undefined \def \showISSN      #1{\unskip}     \fi
\ifx \showLCCN     \undefined \def \showLCCN      #1{\unskip}     \fi
\ifx \shownote     \undefined \def \shownote      #1{#1}          \fi
\ifx \showarticletitle \undefined \def \showarticletitle #1{#1}   \fi
\ifx \showURL      \undefined \def \showURL       {\relax}        \fi
\providecommand\bibfield[2]{#2}
\providecommand\bibinfo[2]{#2}
\providecommand\natexlab[1]{#1}
\providecommand\showeprint[2][]{arXiv:#2}

\bibitem[\protect\citeauthoryear{Bollacker, Evans, Paritosh, Sturge, and Taylor}{Bollacker et~al\mbox{.}}{2008}]%
        {DBLP:conf/sigmod/BollackerEPST08}
\bibfield{author}{\bibinfo{person}{Kurt~D. Bollacker}, \bibinfo{person}{Colin Evans}, \bibinfo{person}{Praveen Paritosh}, \bibinfo{person}{Tim Sturge}, {and} \bibinfo{person}{Jamie Taylor}.} \bibinfo{year}{2008}\natexlab{}.
\newblock \showarticletitle{Freebase: a collaboratively created graph database for structuring human knowledge}. In \bibinfo{booktitle}{\emph{SIGMOD}}. \bibinfo{pages}{1247--1250}.
\newblock
\urldef\tempurl%
\url{https://doi.org/10.1145/1376616.1376746}
\showDOI{\tempurl}


\bibitem[\protect\citeauthoryear{Bordes, Usunier, Garc{\'{\i}}a{-}Dur{\'{a}}n, Weston, and Yakhnenko}{Bordes et~al\mbox{.}}{2013}]%
        {DBLP:conf/nips/BordesUGWY13}
\bibfield{author}{\bibinfo{person}{Antoine Bordes}, \bibinfo{person}{Nicolas Usunier}, \bibinfo{person}{Alberto Garc{\'{\i}}a{-}Dur{\'{a}}n}, \bibinfo{person}{Jason Weston}, {and} \bibinfo{person}{Oksana Yakhnenko}.} \bibinfo{year}{2013}\natexlab{}.
\newblock \showarticletitle{Translating Embeddings for Modeling Multi-relational Data}. \bibinfo{pages}{2787--2795}.
\newblock
\urldef\tempurl%
\url{http://papers.nips.cc/paper/5071-translating-embeddings-for-modeling-multi-relational-data}
\showURL{%
\tempurl}


\bibitem[\protect\citeauthoryear{Ding, Wang, Wang, and Guo}{Ding et~al\mbox{.}}{2018}]%
        {DBLP:conf/acl/WangWGD18}
\bibfield{author}{\bibinfo{person}{Boyang Ding}, \bibinfo{person}{Quan Wang}, \bibinfo{person}{Bin Wang}, {and} \bibinfo{person}{Li Guo}.} \bibinfo{year}{2018}\natexlab{}.
\newblock \showarticletitle{Improving Knowledge Graph Embedding Using Simple Constraints}. In \bibinfo{booktitle}{\emph{ACL}}. \bibinfo{pages}{110--121}.
\newblock
\urldef\tempurl%
\url{https://doi.org/10.18653/v1/P18-1011}
\showDOI{\tempurl}


\bibitem[\protect\citeauthoryear{Dong, Gabrilovich, Heitz, Horn, Lao, Murphy, Strohmann, Sun, and Zhang}{Dong et~al\mbox{.}}{2014}]%
        {DBLP:conf/kdd/0001GHHLMSSZ14}
\bibfield{author}{\bibinfo{person}{Xin Dong}, \bibinfo{person}{Evgeniy Gabrilovich}, \bibinfo{person}{Geremy Heitz}, \bibinfo{person}{Wilko Horn}, \bibinfo{person}{Ni Lao}, \bibinfo{person}{Kevin Murphy}, \bibinfo{person}{Thomas Strohmann}, \bibinfo{person}{Shaohua Sun}, {and} \bibinfo{person}{Wei Zhang}.} \bibinfo{year}{2014}\natexlab{}.
\newblock \showarticletitle{Knowledge vault: a web-scale approach to probabilistic knowledge fusion}. In \bibinfo{booktitle}{\emph{SIGKDD}}. \bibinfo{pages}{601--610}.
\newblock
\urldef\tempurl%
\url{https://doi.org/10.1145/2623330.2623623}
\showDOI{\tempurl}


\bibitem[\protect\citeauthoryear{Gal{\'{a}}rraga, Teflioudi, Hose, and Suchanek}{Gal{\'{a}}rraga et~al\mbox{.}}{2015}]%
        {DBLP:journals/vldb/GalarragaTHS15}
\bibfield{author}{\bibinfo{person}{Luis Gal{\'{a}}rraga}, \bibinfo{person}{Christina Teflioudi}, \bibinfo{person}{Katja Hose}, {and} \bibinfo{person}{Fabian~M. Suchanek}.} \bibinfo{year}{2015}\natexlab{}.
\newblock \showarticletitle{Fast rule mining in ontological knowledge bases with {AMIE+}}.
\newblock \bibinfo{journal}{\emph{{VLDB} J.}} \bibinfo{volume}{24}, \bibinfo{number}{6} (\bibinfo{year}{2015}), \bibinfo{pages}{707--730}.
\newblock
\urldef\tempurl%
\url{https://doi.org/10.1007/s00778-015-0394-1}
\showDOI{\tempurl}


\bibitem[\protect\citeauthoryear{Guo, Wang, Wang, Wang, and Guo}{Guo et~al\mbox{.}}{2018}]%
        {DBLP:conf/aaai/GuoWWWG18}
\bibfield{author}{\bibinfo{person}{Shu Guo}, \bibinfo{person}{Quan Wang}, \bibinfo{person}{Lihong Wang}, \bibinfo{person}{Bin Wang}, {and} \bibinfo{person}{Li Guo}.} \bibinfo{year}{2018}\natexlab{}.
\newblock \showarticletitle{Knowledge Graph Embedding With Iterative Guidance From Soft Rules}. In \bibinfo{booktitle}{\emph{AAAI}}. \bibinfo{pages}{4816--4823}.
\newblock
\urldef\tempurl%
\url{https://www.aaai.org/ocs/index.php/AAAI/AAAI18/paper/view/16369}
\showURL{%
\tempurl}


\bibitem[\protect\citeauthoryear{H{\'{a}}jek}{H{\'{a}}jek}{1998}]%
        {DBLP:books/kl/Hajek98}
\bibfield{author}{\bibinfo{person}{Petr H{\'{a}}jek}.} \bibinfo{year}{1998}\natexlab{}.
\newblock \bibinfo{booktitle}{\emph{Metamathematics of Fuzzy Logic}}. \bibinfo{series}{Trends in Logic}, Vol.~\bibinfo{volume}{4}.
\newblock \bibinfo{publisher}{Kluwer}.
\newblock
\showISBNx{978-1-4020-0370-7}
\urldef\tempurl%
\url{https://doi.org/10.1007/978-94-011-5300-3}
\showDOI{\tempurl}


\bibitem[\protect\citeauthoryear{Hamaguchi, Oiwa, Shimbo, and Matsumoto}{Hamaguchi et~al\mbox{.}}{2017}]%
        {DBLP:conf/ijcai/HamaguchiOSM17}
\bibfield{author}{\bibinfo{person}{Takuo Hamaguchi}, \bibinfo{person}{Hidekazu Oiwa}, \bibinfo{person}{Masashi Shimbo}, {and} \bibinfo{person}{Yuji Matsumoto}.} \bibinfo{year}{2017}\natexlab{}.
\newblock \showarticletitle{Knowledge Transfer for Out-of-Knowledge-Base Entities : {A} Graph Neural Network Approach}. In \bibinfo{booktitle}{\emph{IJCAI}}. \bibinfo{pages}{1802--1808}.
\newblock
\urldef\tempurl%
\url{https://doi.org/10.24963/ijcai.2017/250}
\showDOI{\tempurl}


\bibitem[\protect\citeauthoryear{Hamilton, Ying, and Leskovec}{Hamilton et~al\mbox{.}}{2017}]%
        {DBLP:conf/nips/HamiltonYL17}
\bibfield{author}{\bibinfo{person}{William~L. Hamilton}, \bibinfo{person}{Zhitao Ying}, {and} \bibinfo{person}{Jure Leskovec}.} \bibinfo{year}{2017}\natexlab{}.
\newblock \showarticletitle{Inductive Representation Learning on Large Graphs}. In \bibinfo{booktitle}{\emph{NIPS}}. \bibinfo{pages}{1024--1034}.
\newblock
\urldef\tempurl%
\url{http://papers.nips.cc/paper/6703-inductive-representation-learning-on-large-graphs}
\showURL{%
\tempurl}


\bibitem[\protect\citeauthoryear{Ji, He, Xu, Liu, and Zhao}{Ji et~al\mbox{.}}{2015}]%
        {DBLP:conf/acl/JiHXL015}
\bibfield{author}{\bibinfo{person}{Guoliang Ji}, \bibinfo{person}{Shizhu He}, \bibinfo{person}{Liheng Xu}, \bibinfo{person}{Kang Liu}, {and} \bibinfo{person}{Jun Zhao}.} \bibinfo{year}{2015}\natexlab{}.
\newblock \showarticletitle{Knowledge Graph Embedding via Dynamic Mapping Matrix}. In \bibinfo{booktitle}{\emph{ACL}}. \bibinfo{pages}{687--696}.
\newblock
\urldef\tempurl%
\url{https://www.aclweb.org/anthology/P15-1067/}
\showURL{%
\tempurl}


\bibitem[\protect\citeauthoryear{Ji, Pan, Cambria, Marttinen, and Yu}{Ji et~al\mbox{.}}{2020}]%
        {ji2020survey}
\bibfield{author}{\bibinfo{person}{Shaoxiong Ji}, \bibinfo{person}{Shirui Pan}, \bibinfo{person}{Erik Cambria}, \bibinfo{person}{Pekka Marttinen}, {and} \bibinfo{person}{Philip~S. Yu}.} \bibinfo{year}{2020}\natexlab{}.
\newblock \showarticletitle{A Survey on Knowledge Graphs: Representation, Acquisition and Applications}.
\newblock \bibinfo{journal}{\emph{CoRR}}  \bibinfo{volume}{abs/2002.00388} (\bibinfo{year}{2020}).
\newblock
\showeprint[arxiv]{2002.00388}
\urldef\tempurl%
\url{https://arxiv.org/abs/2002.00388}
\showURL{%
\tempurl}


\bibitem[\protect\citeauthoryear{Kingma and Ba}{Kingma and Ba}{2015}]%
        {DBLP:journals/corr/KingmaB14}
\bibfield{author}{\bibinfo{person}{Diederik~P. Kingma} {and} \bibinfo{person}{Jimmy Ba}.} \bibinfo{year}{2015}\natexlab{}.
\newblock \showarticletitle{Adam: {A} Method for Stochastic Optimization}. In \bibinfo{booktitle}{\emph{ICLR}}, \bibfield{editor}{\bibinfo{person}{Yoshua Bengio} {and} \bibinfo{person}{Yann LeCun}} (Eds.).
\newblock
\urldef\tempurl%
\url{http://arxiv.org/abs/1412.6980}
\showURL{%
\tempurl}


\bibitem[\protect\citeauthoryear{Li}{Li}{2012}]%
        {browne2009jboss}
\bibfield{author}{\bibinfo{person}{Dingcheng Li}.} \bibinfo{year}{2012}\natexlab{}.
\newblock \showarticletitle{Applying JBoss{\textregistered} Drools Business Rules Management System for Electronic Health Records Driven Phenotyping}.
\newblock
\urldef\tempurl%
\url{http://knowledge.amia.org/amia-55142-a2012a-1.636547/t-006-1.640361/f-001-1.640362/a-245-1.640478/a-246-1.640475}
\showURL{%
\tempurl}


\bibitem[\protect\citeauthoryear{Neelakantan, Roth, and McCallum}{Neelakantan et~al\mbox{.}}{2015}]%
        {DBLP:conf/acl/NeelakantanRM15}
\bibfield{author}{\bibinfo{person}{Arvind Neelakantan}, \bibinfo{person}{Benjamin Roth}, {and} \bibinfo{person}{Andrew McCallum}.} \bibinfo{year}{2015}\natexlab{}.
\newblock \showarticletitle{Compositional Vector Space Models for Knowledge Base Completion}. In \bibinfo{booktitle}{\emph{ACL}}. \bibinfo{pages}{156--166}.
\newblock
\urldef\tempurl%
\url{https://www.aclweb.org/anthology/P15-1016/}
\showURL{%
\tempurl}


\bibitem[\protect\citeauthoryear{Nickel, Tresp, and Kriegel}{Nickel et~al\mbox{.}}{2011}]%
        {DBLP:conf/icml/NickelTK11}
\bibfield{author}{\bibinfo{person}{Maximilian Nickel}, \bibinfo{person}{Volker Tresp}, {and} \bibinfo{person}{Hans{-}Peter Kriegel}.} \bibinfo{year}{2011}\natexlab{}.
\newblock \showarticletitle{A Three-Way Model for Collective Learning on Multi-Relational Data}. In \bibinfo{booktitle}{\emph{Proceedings of the 28th International Conference on Machine Learning, {ICML} 2011, Bellevue, Washington, USA, June 28 - July 2, 2011}}. \bibinfo{pages}{809--816}.
\newblock
\urldef\tempurl%
\url{https://icml.cc/2011/papers/438\_icmlpaper.pdf}
\showURL{%
\tempurl}


\bibitem[\protect\citeauthoryear{Omran, Wang, and Wang}{Omran et~al\mbox{.}}{2018}]%
        {omran2019embedding}
\bibfield{author}{\bibinfo{person}{Pouya~Ghiasnezhad Omran}, \bibinfo{person}{Kewen Wang}, {and} \bibinfo{person}{Zhe Wang}.} \bibinfo{year}{2018}\natexlab{}.
\newblock \showarticletitle{Scalable Rule Learning via Learning Representation}. In \bibinfo{booktitle}{\emph{Proceedings of the Twenty-Seventh International Joint Conference on Artificial Intelligence, {IJCAI} 2018, July 13-19, 2018, Stockholm, Sweden}}. \bibinfo{publisher}{ijcai.org}, \bibinfo{pages}{2149--2155}.
\newblock
\urldef\tempurl%
\url{https://doi.org/10.24963/ijcai.2018/297}
\showDOI{\tempurl}


\bibitem[\protect\citeauthoryear{Pritsker, Cohen, and Minkov}{Pritsker et~al\mbox{.}}{2015}]%
        {DBLP:conf/emnlp/PritskerCM15}
\bibfield{author}{\bibinfo{person}{Evgenia~Wasserman Pritsker}, \bibinfo{person}{William~W. Cohen}, {and} \bibinfo{person}{Einat Minkov}.} \bibinfo{year}{2015}\natexlab{}.
\newblock \showarticletitle{Learning to Identify the Best Contexts for Knowledge-based {WSD}}. In \bibinfo{booktitle}{\emph{EMNLP}}. \bibinfo{pages}{1662--1667}.
\newblock
\urldef\tempurl%
\url{https://www.aclweb.org/anthology/D15-1192/}
\showURL{%
\tempurl}


\bibitem[\protect\citeauthoryear{Shang, Tang, Huang, Bi, He, and Zhou}{Shang et~al\mbox{.}}{2019}]%
        {DBLP:conf/aaai/ShangTHBHZ19}
\bibfield{author}{\bibinfo{person}{Chao Shang}, \bibinfo{person}{Yun Tang}, \bibinfo{person}{Jing Huang}, \bibinfo{person}{Jinbo Bi}, \bibinfo{person}{Xiaodong He}, {and} \bibinfo{person}{Bowen Zhou}.} \bibinfo{year}{2019}\natexlab{}.
\newblock \showarticletitle{End-to-End Structure-Aware Convolutional Networks for Knowledge Base Completion}. In \bibinfo{booktitle}{\emph{AAAI}}. \bibinfo{pages}{3060--3067}.
\newblock
\urldef\tempurl%
\url{https://doi.org/10.1609/aaai.v33i01.33013060}
\showDOI{\tempurl}


\bibitem[\protect\citeauthoryear{Shi and Weninger}{Shi and Weninger}{2018}]%
        {DBLP:conf/aaai/ShiW18}
\bibfield{author}{\bibinfo{person}{Baoxu Shi} {and} \bibinfo{person}{Tim Weninger}.} \bibinfo{year}{2018}\natexlab{}.
\newblock \showarticletitle{Open-World Knowledge Graph Completion}. In \bibinfo{booktitle}{\emph{AAAI}}. \bibinfo{pages}{1957--1964}.
\newblock
\urldef\tempurl%
\url{https://www.aaai.org/ocs/index.php/AAAI/AAAI18/paper/view/16055}
\showURL{%
\tempurl}


\bibitem[\protect\citeauthoryear{Socher, Chen, Manning, and Ng}{Socher et~al\mbox{.}}{2013}]%
        {socher2013reasoning}
\bibfield{author}{\bibinfo{person}{Richard Socher}, \bibinfo{person}{Danqi Chen}, \bibinfo{person}{Christopher~D. Manning}, {and} \bibinfo{person}{Andrew~Y. Ng}.} \bibinfo{year}{2013}\natexlab{}.
\newblock \showarticletitle{Reasoning With Neural Tensor Networks for Knowledge Base Completion}. \bibinfo{pages}{926--934}.
\newblock


\bibitem[\protect\citeauthoryear{Sun, Han, Yan, Yu, and Wu}{Sun et~al\mbox{.}}{2011}]%
        {sun2011pathsim}
\bibfield{author}{\bibinfo{person}{Yizhou Sun}, \bibinfo{person}{Jiawei Han}, \bibinfo{person}{Xifeng Yan}, \bibinfo{person}{Philip~S. Yu}, {and} \bibinfo{person}{Tianyi Wu}.} \bibinfo{year}{2011}\natexlab{}.
\newblock \showarticletitle{PathSim: Meta Path-Based Top-K Similarity Search in Heterogeneous Information Networks}.
\newblock \bibinfo{journal}{\emph{Proc. {VLDB} Endow.}} \bibinfo{volume}{4}, \bibinfo{number}{11} (\bibinfo{year}{2011}), \bibinfo{pages}{992--1003}.
\newblock
\urldef\tempurl%
\url{http://www.vldb.org/pvldb/vol4/p992-sun.pdf}
\showURL{%
\tempurl}


\bibitem[\protect\citeauthoryear{Trouillon, Welbl, Riedel, Gaussier, and Bouchard}{Trouillon et~al\mbox{.}}{2016}]%
        {DBLP:conf/icml/TrouillonWRGB16}
\bibfield{author}{\bibinfo{person}{Th{\'{e}}o Trouillon}, \bibinfo{person}{Johannes Welbl}, \bibinfo{person}{Sebastian Riedel}, \bibinfo{person}{{\'{E}}ric Gaussier}, {and} \bibinfo{person}{Guillaume Bouchard}.} \bibinfo{year}{2016}\natexlab{}.
\newblock \showarticletitle{Complex Embeddings for Simple Link Prediction}. In \bibinfo{booktitle}{\emph{ICML}}. \bibinfo{pages}{2071--2080}.
\newblock
\urldef\tempurl%
\url{http://proceedings.mlr.press/v48/trouillon16.html}
\showURL{%
\tempurl}


\bibitem[\protect\citeauthoryear{Wang, Han, Li, and Pan}{Wang et~al\mbox{.}}{2019a}]%
        {DBLP:conf/aaai/WangHLP19}
\bibfield{author}{\bibinfo{person}{PeiFeng Wang}, \bibinfo{person}{Jialong Han}, \bibinfo{person}{Chenliang Li}, {and} \bibinfo{person}{Rong Pan}.} \bibinfo{year}{2019}\natexlab{a}.
\newblock \showarticletitle{Logic Attention Based Neighborhood Aggregation for Inductive Knowledge Graph Embedding}. In \bibinfo{booktitle}{\emph{AAAI}}. \bibinfo{pages}{7152--7159}.
\newblock
\urldef\tempurl%
\url{https://doi.org/10.1609/aaai.v33i01.33017152}
\showDOI{\tempurl}


\bibitem[\protect\citeauthoryear{Wang, Mao, Wang, and Guo}{Wang et~al\mbox{.}}{2017}]%
        {DBLP:journals/tkde/WangMWG17}
\bibfield{author}{\bibinfo{person}{Quan Wang}, \bibinfo{person}{Zhendong Mao}, \bibinfo{person}{Bin Wang}, {and} \bibinfo{person}{Li Guo}.} \bibinfo{year}{2017}\natexlab{}.
\newblock \showarticletitle{Knowledge Graph Embedding: {A} Survey of Approaches and Applications}.
\newblock \bibinfo{journal}{\emph{{IEEE} TKDE}} \bibinfo{volume}{29}, \bibinfo{number}{12} (\bibinfo{year}{2017}), \bibinfo{pages}{2724--2743}.
\newblock
\urldef\tempurl%
\url{https://doi.org/10.1109/TKDE.2017.2754499}
\showDOI{\tempurl}


\bibitem[\protect\citeauthoryear{Wang, Ren, He, Zhang, and Hu}{Wang et~al\mbox{.}}{2019b}]%
        {DBLP:conf/ijcai/WangRHZH19}
\bibfield{author}{\bibinfo{person}{Zihan Wang}, \bibinfo{person}{Zhaochun Ren}, \bibinfo{person}{Chunyu He}, \bibinfo{person}{Peng Zhang}, {and} \bibinfo{person}{Yue Hu}.} \bibinfo{year}{2019}\natexlab{b}.
\newblock \showarticletitle{Robust Embedding with Multi-Level Structures for Link Prediction}. In \bibinfo{booktitle}{\emph{IJCAI}}. \bibinfo{pages}{5240--5246}.
\newblock
\urldef\tempurl%
\url{https://doi.org/10.24963/ijcai.2019/728}
\showDOI{\tempurl}


\bibitem[\protect\citeauthoryear{Wang, Zhang, Feng, and Chen}{Wang et~al\mbox{.}}{2014}]%
        {DBLP:conf/aaai/WangZFC14}
\bibfield{author}{\bibinfo{person}{Zhen Wang}, \bibinfo{person}{Jianwen Zhang}, \bibinfo{person}{Jianlin Feng}, {and} \bibinfo{person}{Zheng Chen}.} \bibinfo{year}{2014}\natexlab{}.
\newblock \showarticletitle{Knowledge Graph Embedding by Translating on Hyperplanes}. In \bibinfo{booktitle}{\emph{AAAI}}. \bibinfo{pages}{1112--1119}.
\newblock
\urldef\tempurl%
\url{http://www.aaai.org/ocs/index.php/AAAI/AAAI14/paper/view/8531}
\showURL{%
\tempurl}


\bibitem[\protect\citeauthoryear{Wu, Khan, Gao, and Li}{Wu et~al\mbox{.}}{2019}]%
        {wu2019efficiently}
\bibfield{author}{\bibinfo{person}{Tianxing Wu}, \bibinfo{person}{Arijit Khan}, \bibinfo{person}{Huan Gao}, {and} \bibinfo{person}{Cheng Li}.} \bibinfo{year}{2019}\natexlab{}.
\newblock \showarticletitle{Efficiently Embedding Dynamic Knowledge Graphs}.
\newblock \bibinfo{journal}{\emph{CoRR}}  \bibinfo{volume}{abs/1910.06708} (\bibinfo{year}{2019}).
\newblock
\showeprint[arxiv]{1910.06708}
\urldef\tempurl%
\url{http://arxiv.org/abs/1910.06708}
\showURL{%
\tempurl}


\bibitem[\protect\citeauthoryear{Xiao, Huang, Meng, and Zhu}{Xiao et~al\mbox{.}}{2017}]%
        {xiao2017ssp}
\bibfield{author}{\bibinfo{person}{Han Xiao}, \bibinfo{person}{Minlie Huang}, \bibinfo{person}{Lian Meng}, {and} \bibinfo{person}{Xiaoyan Zhu}.} \bibinfo{year}{2017}\natexlab{}.
\newblock \showarticletitle{{SSP:} Semantic Space Projection for Knowledge Graph Embedding with Text Descriptions}. In \bibinfo{booktitle}{\emph{Proceedings of the Thirty-First {AAAI} Conference on Artificial Intelligence, February 4-9, 2017, San Francisco, California, {USA}}}. \bibinfo{pages}{3104--3110}.
\newblock
\urldef\tempurl%
\url{http://aaai.org/ocs/index.php/AAAI/AAAI17/paper/view/14306}
\showURL{%
\tempurl}


\bibitem[\protect\citeauthoryear{Xie, Liu, Luan, and Sun}{Xie et~al\mbox{.}}{2017}]%
        {DBLP:conf/ijcai/XieLLS17}
\bibfield{author}{\bibinfo{person}{Ruobing Xie}, \bibinfo{person}{Zhiyuan Liu}, \bibinfo{person}{Huanbo Luan}, {and} \bibinfo{person}{Maosong Sun}.} \bibinfo{year}{2017}\natexlab{}.
\newblock \showarticletitle{Image-embodied Knowledge Representation Learning}. In \bibinfo{booktitle}{\emph{IJCAI}}. \bibinfo{pages}{3140--3146}.
\newblock
\urldef\tempurl%
\url{https://doi.org/10.24963/ijcai.2017/438}
\showDOI{\tempurl}


\bibitem[\protect\citeauthoryear{Xu, Hu, Leskovec, and Jegelka}{Xu et~al\mbox{.}}{2019}]%
        {DBLP:conf/iclr/XuHLJ19}
\bibfield{author}{\bibinfo{person}{Keyulu Xu}, \bibinfo{person}{Weihua Hu}, \bibinfo{person}{Jure Leskovec}, {and} \bibinfo{person}{Stefanie Jegelka}.} \bibinfo{year}{2019}\natexlab{}.
\newblock \showarticletitle{How Powerful are Graph Neural Networks?}. In \bibinfo{booktitle}{\emph{ICLR}}.
\newblock
\urldef\tempurl%
\url{https://openreview.net/forum?id=ryGs6iA5Km}
\showURL{%
\tempurl}


\bibitem[\protect\citeauthoryear{Yang and Mitchell}{Yang and Mitchell}{2017}]%
        {DBLP:conf/acl/YangM17}
\bibfield{author}{\bibinfo{person}{Bishan Yang} {and} \bibinfo{person}{Tom~M. Mitchell}.} \bibinfo{year}{2017}\natexlab{}.
\newblock \showarticletitle{Leveraging Knowledge Bases in LSTMs for Improving Machine Reading}. In \bibinfo{booktitle}{\emph{ACL}}. \bibinfo{pages}{1436--1446}.
\newblock
\urldef\tempurl%
\url{https://doi.org/10.18653/v1/P17-1132}
\showDOI{\tempurl}


\bibitem[\protect\citeauthoryear{Yang, Yih, He, Gao, and Deng}{Yang et~al\mbox{.}}{2015}]%
        {yang2014embedding}
\bibfield{author}{\bibinfo{person}{Bishan Yang}, \bibinfo{person}{Wen{-}tau Yih}, \bibinfo{person}{Xiaodong He}, \bibinfo{person}{Jianfeng Gao}, {and} \bibinfo{person}{Li Deng}.} \bibinfo{year}{2015}\natexlab{}.
\newblock \showarticletitle{Embedding Entities and Relations for Learning and Inference in Knowledge Bases}. In \bibinfo{booktitle}{\emph{3rd International Conference on Learning Representations, {ICLR} 2015, San Diego, CA, USA, May 7-9, 2015, Conference Track Proceedings}}.
\newblock
\urldef\tempurl%
\url{http://arxiv.org/abs/1412.6575}
\showURL{%
\tempurl}


\bibitem[\protect\citeauthoryear{Zhang, Paudel, Wang, Chen, Zhu, Zhang, Bernstein, and Chen}{Zhang et~al\mbox{.}}{2019}]%
        {zhang2019iteratively}
\bibfield{author}{\bibinfo{person}{Wen Zhang}, \bibinfo{person}{Bibek Paudel}, \bibinfo{person}{Liang Wang}, \bibinfo{person}{Jiaoyan Chen}, \bibinfo{person}{Hai Zhu}, \bibinfo{person}{Wei Zhang}, \bibinfo{person}{Abraham Bernstein}, {and} \bibinfo{person}{Huajun Chen}.} \bibinfo{year}{2019}\natexlab{}.
\newblock \showarticletitle{Iteratively Learning Embeddings and Rules for Knowledge Graph Reasoning}. In \bibinfo{booktitle}{\emph{The World Wide Web Conference, {WWW} 2019, San Francisco, CA, USA, May 13-17, 2019}}. \bibinfo{pages}{2366--2377}.
\newblock
\urldef\tempurl%
\url{https://doi.org/10.1145/3308558.3313612}
\showDOI{\tempurl}


\bibitem[\protect\citeauthoryear{Zhang, Zhuang, Qu, Lin, and He}{Zhang et~al\mbox{.}}{2018}]%
        {zhang2018knowledge}
\bibfield{author}{\bibinfo{person}{Zhao Zhang}, \bibinfo{person}{Fuzhen Zhuang}, \bibinfo{person}{Meng Qu}, \bibinfo{person}{Fen Lin}, {and} \bibinfo{person}{Qing He}.} \bibinfo{year}{2018}\natexlab{}.
\newblock \showarticletitle{Knowledge Graph Embedding with Hierarchical Relation Structure}. In \bibinfo{booktitle}{\emph{Proceedings of the 2018 Conference on Empirical Methods in Natural Language Processing, Brussels, Belgium, October 31 - November 4, 2018}}. \bibinfo{pages}{3198--3207}.
\newblock
\urldef\tempurl%
\url{https://doi.org/10.18653/v1/d18-1358}
\showDOI{\tempurl}


\bibitem[\protect\citeauthoryear{Zuo, Fang, Qian, Zhang, and Xu}{Zuo et~al\mbox{.}}{2018}]%
        {xie2016representation}
\bibfield{author}{\bibinfo{person}{Yukun Zuo}, \bibinfo{person}{Quan Fang}, \bibinfo{person}{Shengsheng Qian}, \bibinfo{person}{Xiaorui Zhang}, {and} \bibinfo{person}{Changsheng Xu}.} \bibinfo{year}{2018}\natexlab{}.
\newblock \showarticletitle{Representation Learning of Knowledge Graphs with Entity Attributes and Multimedia Descriptions}. In \bibinfo{booktitle}{\emph{Fourth {IEEE} International Conference on Multimedia Big Data, BigMM 2018, Xi'an, China, September 13-16, 2018}}. \bibinfo{pages}{1--5}.
\newblock
\urldef\tempurl%
\url{https://doi.org/10.1109/BigMM.2018.8499179}
\showDOI{\tempurl}


\end{thebibliography}
\end{document}